\documentclass[sigconf]{acmart}

\usepackage{booktabs} 

\usepackage{graphicx} 
\usepackage{amsmath}
\usepackage{comment}
\usepackage{amsfonts}
\usepackage{xcolor}
\usepackage{multirow}
\usepackage{multicol}
\usepackage{subfigure}
\usepackage{algorithm}  
\usepackage{algorithmic}
\usepackage{array}
\usepackage{caption}
\usepackage{enumitem}
\usepackage{footnote}

\copyrightyear{2019}
\acmYear{2019} 

\usepackage{balance}

\begin{document}

\title{Iteratively Learning Embeddings and Rules for Knowledge Graph Reasoning}

\author{Wen Zhang}
\affiliation{%
  \institution{College of Computer Science, Zhejiang University}
  \institution{}
}
\email{wenzhang2015@zju.edu.cn}

\author{Bibek Paudel}
\authornote{Work done while at UZH.}
\affiliation{%
  \institution{Stanford University \\ \& University of Z\"urich}
}
\email{paudel@ifi.uzh.ch}

\author{Liang Wang}
\affiliation{%
  \institution{College of Computer Science, Zhejiang University}
}
\email{21621254@zju.edu.cn}

\author{Jiaoyan Chen}
\affiliation{%
  \institution{Department of Computer Science,  University of Oxford}
}
\email{jiaoyan.chen@cs.ox.ac.uk}

\author{Hai Zhu}
\affiliation{%
  \institution{Alibaba Group}
  \country{China}
}
\email{marvin.zh@alibaba-inc.com}

\author{Wei Zhang}
\affiliation{%
  \institution{Alibaba Group \& AZFT Joint Lab for Knowledge Engine}
}
\email{lantu.zw@alibaba-inc.com}

\author{Abraham Bernstein}
\affiliation{%
  \institution{Department of Informatics,\; University of Z\"urich}
  \country{Switzerland}
}
\email{bernstein@ifi.uzh.ch}

\author{Huajun Chen}
\authornote{Corresponding author.}
\affiliation{%
  \institution{Zhejiang University \& AZFT Joint Lab for Knowledge Engine}
  \country{China}
}
\email{huajunsir@zju.edu.cn}


\begin{abstract}
Reasoning is essential for the development of large knowledge graphs, especially for completion, which aims to infer new triples based on existing ones. Both rules and embeddings can be used for knowledge graph reasoning and they have their own advantages and difficulties. Rule-based reasoning is accurate and explainable but rule learning with searching over the graph always suffers from efficiency due to huge search space. Embedding-based reasoning is more scalable and efficient as the reasoning is conducted via computation between embeddings, but it has difficulty learning good representations for sparse entities because a good embedding relies heavily on data richness. Based on this observation, in this paper we explore how embedding and rule learning can be combined together and complement each other's difficulties with their advantages. We propose a novel framework IterE iteratively learning embeddings and rules, in which rules are learned from embeddings with proper pruning strategy and embeddings are learned from existing triples and new triples inferred by rules. Evaluations on embedding qualities of IterE show that rules help improve the quality of sparse entity embeddings and their link prediction results. We also evaluate the efficiency of rule learning and quality of rules from IterE compared with AMIE+, showing that IterE is capable of generating high quality rules more efficiently. Experiments show that iteratively learning embeddings and rules benefit each other during learning and prediction.
\end{abstract}

\vspace{-2mm}

\maketitle


\vspace{-2.5mm}
\section{Introduction}
\label{sec:introduction}
Many Knowledge Graphs (KGs), such as Freebase~\cite{Freebase:conf/sigmod/BollackerEPST08} and YAGO~\cite{YAGO:conf/www/SuchanekKW07}, have been built in recent years and led to a broad range of applications, including question answering~\cite{KBQA:journals/pvldb/CuiXWSHW17}, relation extraction~\cite{LFDS:guanyingEMNLP2018}, and recommender system~\cite{CKE:conf/kdd/ZhangYLXM16}.
KGs store facts as triples in the form of 
\emph{(subject entity, relation, object entity)}, abridged as $(s, r, o)$. Some KGs also have an ontology with class and property expression axioms which place constraints on classes and types of relationships. 

Knowledge graph reasoning (KGR) can infer new knowledge based on existing ones and check knowledge consistency. It is attracting research interest and is important for completing and cleaning up KGs. 
Two of the most common learning methods for KGR are embedding-based reasoning and rule-based reasoning ~\cite{review:journals/pieee/Nickel0TG16}.

One of the crucial tasks for embedding-based and rule-based reasoning is to learn embeddings and rules respectively. 
\emph{Embedding learning methods} such as TransE~\cite{TransE:conf/nips/BordesUGWY13}, HolE~\cite{HolE:conf/aaai/NickelRP16} and ComplEx~\cite{ComplEx:conf/icml/TrouillonWRGB16} learn latent representations of entities and relations in continuous vector spaces, called \emph{embeddings}, so as to preserve the information and semantics in KGs. Embedding-based reasoning is more efficient when there are a large number of relations or triples to reason over. 
\emph{Rule learning methods} such as AMIE\cite{AMIE:conf/www/GalarragaTHS13} aim to learn
deductive and interpretable inference rules. Rule-based reasoning is precise and can provide insights for inference results.

With different advantages, both embedding and rule learning are important for KGR, while they still have their own difficulties and weaknesses.

\begin{figure}[!hbpt]
\centering 
\vspace{-3mm}
\includegraphics[scale=0.60]{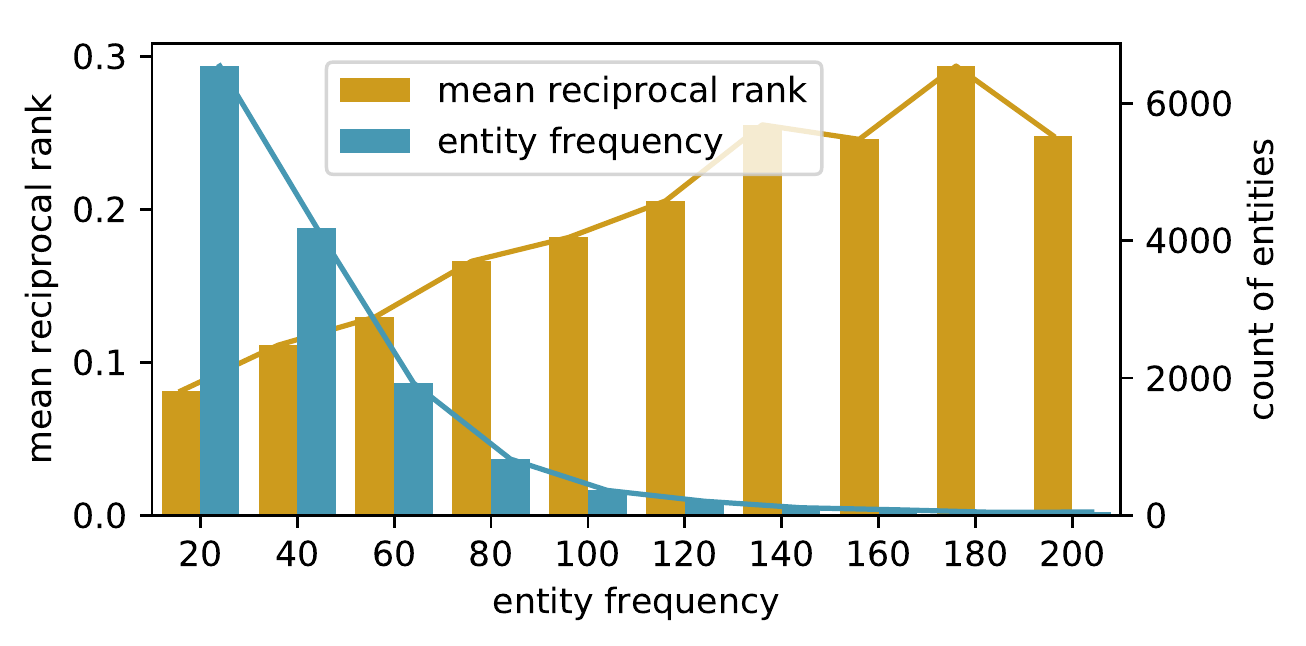}
\vspace{-4mm}
\caption{
\small{The numbers and mean reciprocal rank of different frequency entities based on ANALOGY results on FB15k-237.}} 
\vspace{-4mm}
\label{sparse-problem}
\end{figure}

\textbf{Sparsity Problem for Embedding Learning.} 
One of the main difficulties for embedding learning is the poor capability of encoding sparse entities. 
For example, Figure~\ref{sparse-problem} shows correlation between entity frequency and entity link prediction results measured in mean reciprocal rank (MRR), 
where higher values means better results.
In Figure \ref{sparse-problem}, the blue line shows there are a large portion of entities having only a few triples, revealing the common existence of sparse entity.  
The yellow line shows that the prediction results of entities  are highly related to their frequency, and the results of sparse entities are much worse than those of frequent ones.

\textbf{Efficiency Problem for Rule Learning.} 
The main difficulty in rule learning is the huge search space when determining rule structures and searching for support triples. 
For example, with a small KG containing $10$ relations and $100$ entities, 
the number of possible structures for a rule with $3$ relations is $10^3$ and the maximum number of supports for these rules is $100^{2*3}*10^3 = 10^{15}$. 
Since the search space is exponential to the number of relations, it will be much larger for real KGs than this example.

With different advantages and difficulties, we argue that embedding learning and rule learning can benefit and complement each other.
On the one hand, deductive rules can infer additional triples for sparse entities and help embedding learning methods encode them better. 
On the other hand, embeddings encoded with rich semantics can turn rule learning from discrete graph search into vector space calculation, so that reduce the search space significantly.
Thus we raise the research question: 
\textbf{whether it is possible to learn embeddings and rules at the same time and make their advantages complement to each other's difficulties.}

In this paper, we propose a novel framework \textbf{IterE} that iteratively learns embeddings and rules, which can combine many embedding methods and different kinds of rules. 
Especially, we consider linear map assumption (Section \ref{sec:preliminaries-KGE}) for embedding learning because it is inherently friendly for rule learning as there are special rule conclusions for relation embeddings in rules (Table~\ref{linear_map_properties}). 
We also consider a particular collection of object property  axioms defined in OWL2 (Table~\ref{tab:axiom-condition-owl2}) for rule learning considering that semantics included in web ontology language are important for the development of knowledge graph. 

IterE mainly includes three parts: (1) embedding learning, (2) axiom induction, and (3) axiom injection. 
Embedding learning learns embeddings for entities and relations, with input including triples existing in KG and those inferred by rules.  
Axiom induction first generates a pool of possible axioms with an effective pruning strategy proposed in this paper and then 
assigns a score to each axiom in the pool based on calculation between relation embeddings according to rule conclusions from linear map assumption. 
Axiom injection utilizes axioms' deductive capability to infer new triples about sparse entities according to axiom groundings and injects these new triples into KG to improve sparse entity embeddings.
These three parts are conducted iteratively during training.

We evaluate IterE from three perspectives, 1) whether axiom improves sparse embeddings' quality and their predictions, 2) whether embedding helps improve rule learning efficiency and quality, and 3) how iterative training improves both embedding and rule learning during training.
The experiment results show that IterE achieves both better link prediction performance and high quality rule learning results. These support our goal of making IterE complement the strengths of embedding and rule learning.

Contributions of our work are as follows:
\begin{itemize}
\vspace{-1mm}
	\item We propose an iterative framework that combines embedding learning and rule learning to explore the mutual benefits between them. Experiments show that it leads to better link prediction results using rules and embeddings together.
	\item We present a new method for \emph{embedding learning with rules} based on axiom injection through t-norm based fuzzy logics. Experiments show that IterE can significantly improve embedding quality for the sparse part of a knowledge graph. 
	\item We further identify a portfolio of ontology axioms for \emph{rule learning with embedding} based on linear map assumption. Experiments show that IterE learns more high quality rules more efficiently than conventional rule learning systems. 
	\end{itemize}

\vspace{-1mm}

\vspace{-3mm}
\section{Preliminaries}
\label{sec:preliminaries}
\begin{table*}[!htb]
\centering
\footnotesize
\caption{ 
\footnotesize{Conditions for object property expression axioms in OWL2 and translated rule formulation. $\mathtt{OP}$ refers to "ObjectProperty".  $\mathtt{OPE}$, with or without subscript, denotes Object Property Expression and $\mathtt{x, y, z}$ are entity variables. $\triangle _I$ is a nonempty set called object domain. $\cdot^\mathtt{OP}$ is an object property interpretation function. When translating axioms into rule forms according to condition, we replace $\mathtt{OPE}$ in axioms with binary relation $r$ in the context of knowledge graph and the number of $\mathtt{OPE}$ in $\mathtt{EquivalentOP}$} and the $\mathtt{OPChain}$ in $\mathtt{SubOP}$ is set to $2$. 
}
\vspace{-4mm}
\begin{tabular}{|l|l| l|}
\hline
\textbf{Object Property Axiom} & \textbf{Condition} & \textbf{Rule Form}\\
\hline
$\mathtt{ReflexiveOP(OPE)}$ 
	& $\forall \mathtt{x: x} \in \triangle _I$ implies (x,x) $\in \mathtt{(OPE)}^{\mathtt{OP}}$
	& $(x, r, x)$ \footnotemark 
	  \\ 
\hline
$\mathtt{SymmetricOP(OPE)}$ 
	& $ \forall \mathtt{x,y: (x, y)} \in \mathtt{(OPE)^{OP}}$ implies $\mathtt{(y,x) \in (OPE) ^{OP}}$ 
	& $(y, r, x) \gets (x, r, y)$\\
	
\hline


$\mathtt{TransitiveOP(OPE)}$ 
	& $\forall \mathtt{x,y,z:(x,y) \in (OPE)^{OP}}$ and $\mathtt{(y,z) \in (OPE)^{OP}}$ imply $\mathtt{(x,z) \in (OPE)^{OP}}$ 
& $(x, r, z) \gets (x, r, y), (y, r, z)$\\

\hline
$\mathtt{EquivalentOP(OPE_1\  ...\ OPE_n)}$ 
	& $\mathtt{(OPE_j)^{OP} = (OPE_k)^{OP}}$ for each $1 \le j \le n$  and each $1 \le k \le n$
	& $(x, r_2, y) \gets (x, r_1, y)$ \\

\hline
$\mathtt{SubOP(OPE_1\ OPE_2)}$
	& $ \mathtt{(OPE_1)^{OP} \subseteq (OPE_2)^{OP}}$  
	& $(x, r_2, y) \gets (x, r_1, y)$ \\ 
\hline

$\mathtt{InverseOP(OPE_1\  OPE_2)}$
	& $\mathtt{(OPE_1)^{OP} = \{(x, y)| (y,x) \in (OPE_2)^{OP} \}}$  
	& $(x, r_1, y) \gets (y, r_2, x)$\\
\hline


\multirow{2}{*}{$\mathtt{SubOP(OPChain(OPE_1\ ...\ OPE_n)\ OPE)}$} 
	& $\forall \mathtt{y_0,...,y_n: (y_0, y_1) \in (OPE)^{OP}}$ and ... and $\mathtt{(y_{n-1}, y_n) \in (OPE_n)^{OP}}$
	& \multirow{2}{*}{$(y_0, r, y_2) \gets (y_0, r_1, y_1),(y_1, r_2, y_2)$}\\
	&imply $\mathtt{(y_0, y_n) \in (OPE)^{OP}}$ 
	&\\

\hline
\end{tabular}

\label{tab:axiom-condition-owl2}
\vspace{-5mm}
\end{table*}

\subsection{Knowledge graph embedding} 
\label{sec:preliminaries-KGE}
A KG $\mathcal{K} = \{\mathcal{E}, \mathcal{R}, \mathcal{T} \}$ contains a set of entities $\mathcal{E}$, a set of relations $\mathcal{R}$ and a set of triples $\mathcal{T} = \{ (s, r, o) | s,o \in \mathcal{E}; r \in \mathcal{R} \}$. In a triple $(s, r, o)$, the symbols $s, r,$ and $o$ denote subject entity, relation, and object entity respectively. An example of such triple is 
$(\mathtt{Tokyo}, locatedIn, \mathtt{Japan})$.
 
Knowledge graph embedding (KGE) aims to embed all entities and relations in a continuous vector space, usually as vectors or matrices called $embeddings$.
Embeddings can be used to estimate the likelihood of a triple to be true via a score function $f: \mathcal{E} \times \mathcal{R} \times \mathcal{E} \to \mathbb{R}$.
Concrete score functions are defined based on different vector space assumptions. 
We now describe two vector space assumptions commonly used in KGEs and their corresponding score functions.

(a) Translation-based assumption embeds entities and relations as vectors and assumes $\textbf{v}_s + \textbf{v}_r = \textbf{v}_o$, in which $\textbf{v}_s, \textbf{v}_r$ and $\textbf{v}_o$ are vector embeddings for $s, r$ and $o$ respectively. For a true triple, the relation-specific translation of subject embedding ($\bf{v}_s + \bf{v}_r$) is close to the object embedding $\bf{v}_o$ in embeddings' vector space.

(b) Linear map assumption embeds entities as vectors and relations as matrices. It assumes that the subject entity embedding $\textbf{v}_s$ can be linearly mapped to object entity embedding $\textbf{v}_o$ via relation embedding $\textbf{M}_r$. In this case, for a true triple, the linear mapping of the subject embedding by the relation matrix ($\bf{v}_s\bf{M}_r$) is close to the object embedding $\bf{v}_o$ in embeddings' vector space.

Thus the score function $\phi$ of two assumptions can be written as: 
\begin{align}
\label{eq:linearmap}
\phi_{translation} = sim(\textbf{v}_s + \textbf{v}_r, \textbf{v}_o)\nonumber \\
\phi_{linearmap} = sim(\textbf{v}_s\textbf{M}_r, \textbf{v}_o) 
\end{align}
where $sim(\bf{x}, \bf{y})$ calculates the similarity between vector $\bf{x}$ and $\bf{y}$.

From the assumption point of view, $\bf{v_s}\bf{M_r} = \bf{v_o}$ (or $\bf{v_s} + \bf{v}_r = \bf{v_o}$) should exactly hold for true triples in the linear-map assumption (or translation-based assumption). From the modeling point of view, this is the optimization goal during learning, namely to be as close as possible to the equation in assumption. The learning process is done through either maximizing the objective or minimizing the error induced from assumptions given by their respective loss functions. Hence, the assumption equation usually does not exactly hold with learned embeddings, but their loss functions are designed to approach the assumption as much as possible. 

In this paper, we adopt linear map assumption for embedding learning because many reasonable rule conclusions can be derived with relation embeddings based on this assumption (Table \ref{linear_map_properties}).

\vspace{-3mm}
\subsection{Rules Learning}

Suppose $\mathcal{X}$ is a countable set of variables and $\mathcal{C}$ a countable set of constants. 
A rule is of the form $head \gets body$, where $head$ is an atom over $\mathcal{R} \cup \mathcal{X} \cup{C}$ and \emph{body} is a conjunction of positive or negative atoms over $\mathcal{R} \cup \mathcal{X} \cup{C}$. An example of such rule can be: 
\begin{equation}
	 (\mathrm{X}, hasMother, \mathrm{Y}) \gets (\mathrm{X},hasParent,\mathrm{Y}), (\mathrm{Y}, gender, \mathrm{Female})
	 \label{rule-example}
\end{equation}
When replacing all variables in a rule with concrete entities in KG, we get a $grounding$ of the rule. For example, one grounding of Rule~(\ref{rule-example}) can be:
\begin{align}
\vspace{-1mm}
 \label{rule-grounding}
	 (\mathrm{Bob},  & hasMother,  \mathrm{Ann}) \gets \\ \nonumber
	 & (\mathrm{Bob}, hasParent, \mathrm{Ann}), (\mathrm{Ann}, gender,\mathrm{Female})
	 \vspace{-1mm}
\end{align}
A grounding with all triples existing in knowledge graph is a $support$ of this rule. For examples, if  $(\mathrm{Bob}, hasMother, \mathrm{Ann}) \in \mathcal{K}$, $(\mathrm{Bob}, hasParent, \mathrm{Ann}) \in \mathcal{K}$ and $(\mathrm{Ann}, gender, \mathrm{Female}) \in \mathcal{K}$, then grounding (\ref{rule-grounding}) is a support for rule (\ref{rule-example}).  

The results of rule learning is of the following form:
\vspace{-1mm}
\begin{equation*}
	\alpha \; head \gets body 
\end{equation*}
\vspace{-1mm}
in which $\alpha \in [0, 1]$ is a confidence score assigned to the rule by the learning method. 

Incorporating logical rules into other learning system such as embedding learning is called \emph{rule injection}. One way for rule injection is adding regularizer or other constraints to entity and relation representations by propositionalizing the rules. Another way is adding constraints to the constituent relations' representation in rules without direct effect on  entity representations. As we want to get new information of sparse entities through rules, we chose propositionalization in this paper for rule injection.

\vspace{-3mm}
\subsection{OWL 2 Web Ontology Language Axioms}
\footnotetext[1]{
the translated rule form of $\mathtt{ReflexiveOP(OPE)}$ only contain a \emph{head}.
}
In this paper, instead of learning general Horn rules or closed-path rules as previous works\cite{AMIE+:journals/vldb/GalarragaTHS15}\cite{NeuralLP:conf/nips/YangYC17}, 
we are more interested in ontology axioms, the main components of knowledge graph ontologies, because they are important for enriching semantics in KGs.

OWL2 Web Ontology Language, informally OWL2\footnote{https://www.w3.org/TR/owl2-primer/}, is an ontology language for Semantic Web with formally defined meaning and is a W3C recommendation.
It defines multiple types of axioms, from which we select some of them as a guidance of rule structures. The selection is based on following principles: (i) the axioms are related with binary relations, the main components of rules in KG and (ii) they can infer new triples because rules are used to help add new information about sparse entities in this paper. 
Thus, we focus on object property expression axioms in OWL2 which are composed of binary relations in the context of KG. Finally, $7$ types of object property expression axioms out of $14$ are selected. The unselected axioms are mainly applied to help check the consistency in knowledge graph.

In OWL2, each axiom has its own condition revealing its semantics. The axiom is \emph{satisfied} if its condition hold. We introduce the selected $7$ types of object property expression axioms and their conditions in Table \ref{tab:axiom-condition-owl2}. We also translate the conditions of axioms into rule-form including a \emph{head} and a \emph{body} as introduced in the previous subsection. The translated rule forms are used to guide the structures of rules to be learned in this paper.

\vspace{-2mm}
\section{Method}
\label{sec:method}
\begin{table*}[!htb]
\centering
\small
\vspace{-3mm}
\caption{ 
\small{Seven types of object property expression axioms selected from OWL2 ontology language. $\mathtt{OP}$ is the short for $\mathtt{ObjectProperty}$.
$\mathcal{K}$ denotes a KG and $x, y, z$ are entity variables. 
 $\textbf{v}$ and $\textbf{M}$ denote entity and relation embeddings respectively.
 $\textbf{I}$ is identity matrix. 
 }
}
\vspace{-4mm}
\begin{tabular}{|c|c|c|c|}
\hline
\textbf{Object Property Axioms} 
	& \textbf{Rule Form}
	& \textbf{According to Linear Map Assumption} 
	& \textbf{Rule Conclusion} \\
\hline
$\mathtt{ReflexiveOP}(r)$ 
	&  $(x, r, x)$
	& $\textbf{v}_x \textbf{M}_r = \textbf{v}_x$ 
	& $\textbf{M}_r = \textbf{I}$ \\
\hline
$\mathtt{SymmetricOP}(r)$ 
	& $(y, r, x) \gets (x, r, y)$
	&  $\textbf{v}_y\textbf{M}_r=\textbf{v}_x$; $\textbf{v}_x\textbf{M}_r=\textbf{v}_y $
	& $ \textbf{M}_r\textbf{M}_r = \textbf{I}$ \\
\hline
$\mathtt{TransitiveOP}(r)$ 
	& $(x, r, z) \gets (x, r, y), (y, r, z)$
	& $\textbf{v}_x\textbf{M}_r=\textbf{v}_z; \textbf{v}_x\textbf{M}_r=\textbf{v}_y,\; \textbf{v}_y\textbf{M}_r=\textbf{v}_z,$
	& $\textbf{M}_r\textbf{M}_r = \textbf{M}_r$ \\
\hline
$\mathtt{EquivalentOP}(r_1, r_2)$ 
	& $(x, r_2, y) \gets (x, r_1, y)$
	& $\textbf{v}_x\textbf{M}_{r_2}=\textbf{v}_y,\; \textbf{v}_x\textbf{M}_{r_1}=\textbf{v	}_y$
	& $\textbf{M}_{r_1}=\textbf{M}_{r_2}$ \\
\hline
$\mathtt{subOP}(r_1, r_2)$
	& $(x, r_2, y) \gets (x, r_1, y)$  
	& $\textbf{v}_x\textbf{M}_{r_2} = \textbf{v}_y,\; \textbf{v}_x\textbf{M}_{r_1} = \textbf{v}_y$
	&  $\textbf{M}_{r_1}=\textbf{M}_{r_2}$  \\ 
\hline
$\mathtt{inverseOP}(r_1, r_2)$
	& $(x, r_1, y) \gets (y, r_2, x)$
	& $\textbf{v}_x\textbf{M}_{r_1} = \textbf{v}_y,\; \textbf{v}_y\textbf{M}_{r_2} = \textbf{v}_x$ 
	&  $\textbf{M}_{r_1}\textbf{M}_{r_2} = \textbf{I}$ \\
\hline
$\mathtt{subOP(OPChain}(r_1, r_2), r)$ 
	& $(y_0, r, y_2) \gets (y_0, r_1, y_1),(y_1, r_2, y_2)$ 
	& $\textbf{v}_{y_0}\textbf{M}_{r}=\textbf{v}_{y_2},\; \textbf{v}_{y_0}\textbf{M}_{r_1}=\textbf{v}_{y_1},\; \textbf{v}_{y_1}\textbf{M}_{r_2}=\textbf{v}_{y_2}$
	& $\textbf{M}_{r_1}\textbf{M}_{r_2} = \textbf{M}_{r}$ \\
\hline
\end{tabular}

\label{linear_map_properties}
\vspace{-4mm}

\end{table*}
Given a knowledge graph $\mathcal{K}= \{ \mathcal{E},\mathcal{R},\mathcal{T} \}$, 
our goal is to learn  embeddings and rules at the same time and make their advantages complement each other's difficulties. As discussed in Section~\ref{sec:introduction}, embedding learning methods suffer from the problem of data sparsity and rule learning methods have a very large search space.

\begin{figure}[!hpt]
\centering 
\includegraphics[scale=0.23]{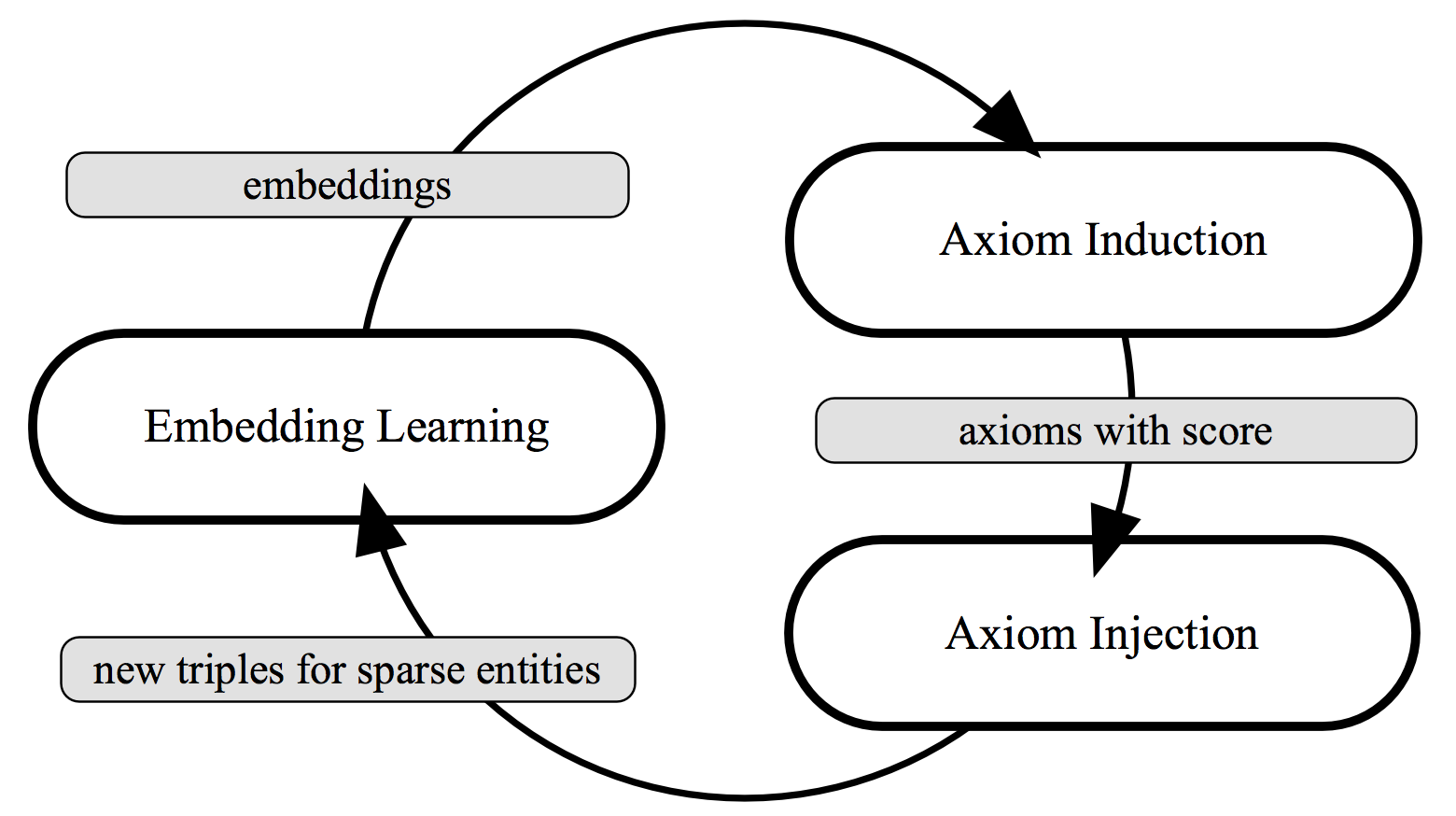}
\caption{\small{Overview of our IterE.}}
\label{basic-idea}
\vspace{-5mm}
\end{figure}

In this paper, we propose a general framework \textbf{IterE} which learns embeddings and rules in an iterative manner and can be applied to many KGEs that are based on linear map assumption. It includes three main parts:(i) embedding learning, (ii) axiom induction and (iii) axiom injection. 
Figure~\ref{basic-idea} shows the key idea of IterE with these three iterative parts.

\begin{itemize}[leftmargin=*, wide]
	\item \textbf{Embedding learning} learns entity embeddings $\mathbf{E}$ and relation embeddings $\mathbf{R}$ with a loss function $L_{embedding}$ to be minimized, calculated with input triples $(s,r,o)$, each with a label related with its truth value. \; The inputs are of two types: triples existing in $\mathcal{K}$ and triples that do not exist in $\mathcal{K}$ but are inferred by axioms.
	\item \textbf{Axiom Induction} inducts a set of axioms $\mathcal{A}$ based on relation embeddings $\textbf{R}$ from the embedding learning step, and assigns each axiom with a score $s_{axiom}$. 
	\item \textbf{Axiom Injection} injects new triples about sparse entities in $\mathcal{K}$ to help improve their poor embeddings caused by insufficient training. The new triples are inferred from groundings of quality axioms with high scores in $\mathcal{A}$ from axiom induction. After axiom injection, with $\mathcal{K}$ updated, the process goes back to embedding learning again.
\end{itemize} 

IterE is proposed based on the observation that embeddings learned with linear map assumption can fully support the axiom selected in this paper, while other assumptions such as translation-based assumption can't as pointed out in ~\cite{ORC:conf/www/Zhang17}.
Inherently, for each type of axioms, a meaningful conclusion can be drawn with relation embeddings according to linear map assumption.
For example, considering axiom $\mathtt{inverse}(hasParent, hasChild)$, if $(Mike, hasParent,$ $John)$ exists in knowledge graph, according to the condition and rule form of inverse axiom in Table \ref{tab:axiom-condition-owl2}, another triple $(John, hasChild,$ $Mike)$ can be inferred. 
Suppose embeddings of $Mike$, $John$, $hasParent$ and $hasChild$ are $\textbf{v}_{Mike}$, $\textbf{v}_{John}$, $\textbf{M}_{hasChild}$ and $\textbf{M}_{hasParent}$ respectively. According to the linear assumption for individual triples, we can get following two equations:
$\textbf{v}_{Mike}\textbf{M}_{hasParent} = \textbf{v}_{John}$ and $\textbf{v}_{John}\textbf{M}_{hasChild} = \textbf{v}_{Mike}$. 
With these two equations, another equation can be deduced: $\textbf{M}_{hasParent}\textbf{M}_{hasChild} = \textbf{I}$. Note that this conclusion equation is only related with  two corresponding relation embeddings and is unrelated with concrete entities, thus it can be regarded as a general conclusion for axiom $\mathtt{inverse}(hasParent, $ $hasChild)$. For other types of axioms, a general conclusion can be drawn in the same way.
We list details of the conclusion for each type of axioms in Table~\ref{linear_map_properties}. 
These conclusions about relation embeddings help guide axiom induction in this paper.

\begin{figure*}[!hpt]
\centering 
\vspace{-3mm}
\includegraphics[scale=0.41]{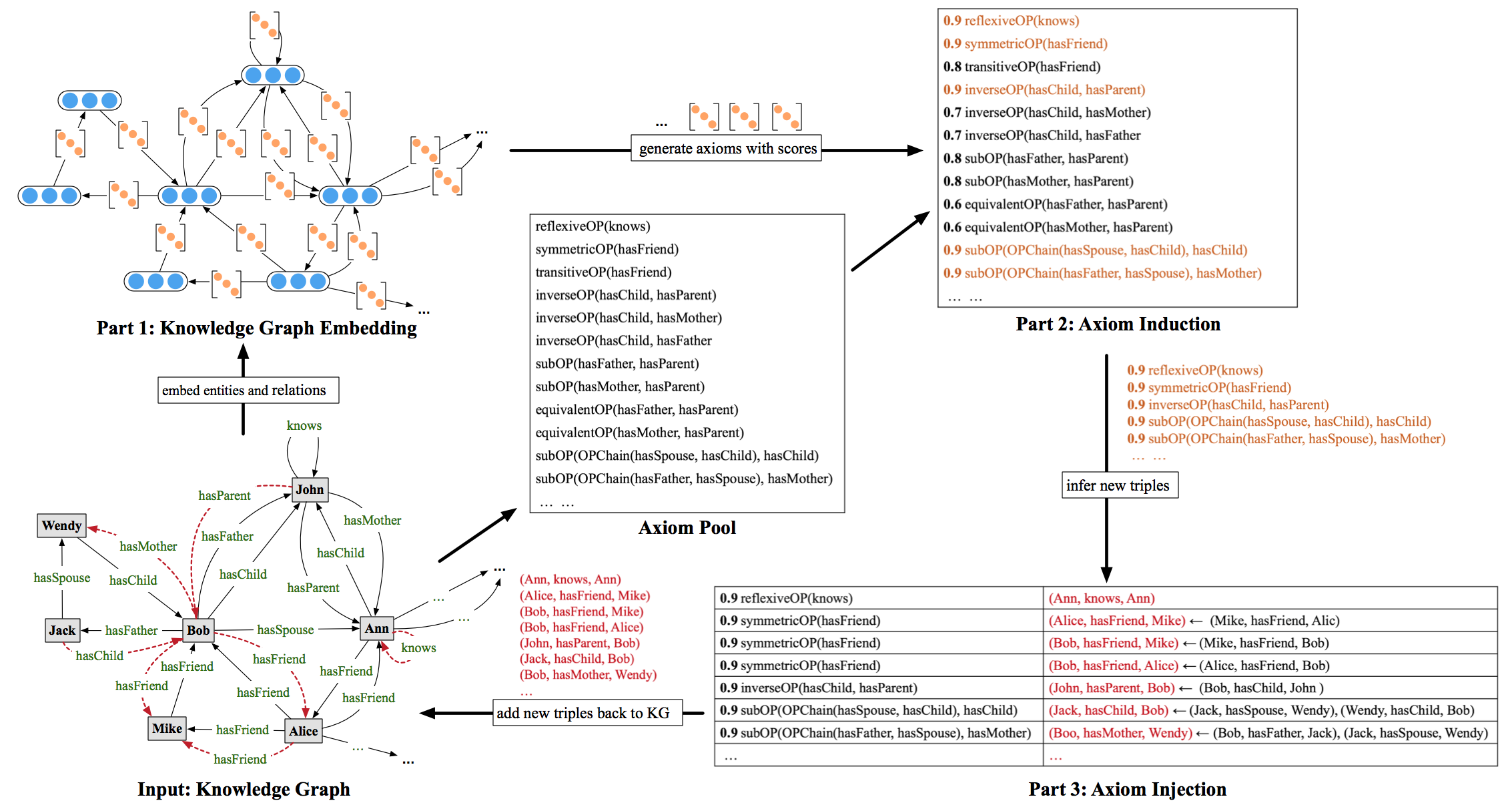}
\vspace{-4mm}
\caption{\small{Detailed example for the process of IterE.}}
\label{basic-idea-examples}
\vspace{-5mm}
\end{figure*}

\vspace{-3mm}
\subsection{Embedding Learning}
The input $\mathcal{I}$ of embedding learning
is a set of  triples with labels. Unlike previous KGEs, which only learn embeddings but not rules, positive input triple in IterE contains two parts: triples $(s, r, o) \in \mathcal{T}$ existing in original knowledge graph  and triples $(s, r, o) \in \mathcal{T}_{axiom}$, where $\mathcal{T}_{axiom}$ is a set of triples inferred by axioms learned from embeddings. 
Negative input triples $(s^\prime, r^\prime, o^\prime) \in \mathcal{T}_{negative}$ are generated by randomly replacing $s$ or $o$ with $e \in \mathcal{E}$ or replacing $r$ with  $r^\prime \in \mathcal{R}$ for $(s, r, o) \in \mathcal{T}$. 
Thus the input set $\mathcal{I}$ in IterE is as follows:
\vspace{-1mm}
\begin{equation}
\vspace{-1mm}
\mathcal{I} = \{((s, r, o), l_{sro}) | ((s, r, o )) \in {\mathcal{T} \land \mathcal{T}_{axiom} \land \mathcal{T}_{negative}} \}
\end{equation}
where $l_{sro}$ is the label for triple $(s, r, o)$ to evaluate its truth value. $l_{sro} = 1$ for $(s, r, o) \in \mathcal{T}$ and $l_{sro}=0$ for $(s, r, o) \in \mathcal{T}_{negative}$. 
For triples $(s, r, o) \in \mathcal{T}_{axiom}$ we assign $l_{sro} = \pi (s, r, o)$,  where $\pi (s, r, o)$ is triple truth value predicted by axiom injection (Section \ref{sec:axiom-injection}).

With input $\mathcal{I}$, the loss function of embedding learning is calculated by mean of cross entropy loss among all $n$ input triples and the training goal is to minimize the following loss function:
\vspace{-1mm}
\begin{align}
\vspace{-1mm}	
\mathrm{min}\; L_{embedding} = \frac{1}{n}&\sum_{((s, r, o), l_{sro}) \in \mathcal{I}} [-l_{sro}\mathrm{log}(\phi(s,r,o)) \notag \\
	&-(1-l_{sro})\mathrm{log}(1-\phi(s,r,o))]
	\label{score funtion2} 
\end{align}

As we adopt linear map assumption, the score function $\phi (s, r, o)$ for each triple $(s, r,o)$ is defined as:
\vspace{-1mm}
\begin{equation}
\vspace{-1mm}
	\phi (s, r, o) = sim(\textbf{v}_s \textbf{M}_r, \textbf{v}_o) =  \sigma(\textbf{v}_s ^\top \textbf{M}_r \textbf{v}_o)
	\label{score-function}
\end{equation}
in which $\textbf{v}_s \in \mathbb{R}^{1 \times d}, \textbf{v}_o \in \mathbb{R}^{1 \times d}$ are vector embeddings for subject and object entity. $\textbf{M}_r \in \mathbb{R}^{d \times d}$ is matrix embedding for relation and $\sigma$ denotes the sigmoid function. $d$ is the embedding dimension. The similarity between two vectors is evaluated via dot product. 

Our approach can be combined with many KGEs based on linear map assumption, such as DistMult\cite{DistMult:conf/iclr/2015} and ANALOGY\cite{ANALOGY:conf/icml/LiuWY17}.
In this paper, we adopt ANALOGY as it achieves state-of-the-art results on link prediction. 
ANALOGY is proposed to deal with an important kind of reasoning, analogy reasoning, in embedding learning. It imposes analogical structures among embeddings and requires linear maps associated with relations to form a commuting family of normal matrices. Specifically, the relation matrix embeddings are constrained as block-diagonal matrices with each diagonal block is either a real scalar, or a 2-dimensional real matrix in the form of 
$\begin{bmatrix}
	a & -b \\
	b & a 
\end{bmatrix}$,
where both $a$ and $b$ are real scalars.

After embedding learning, a collection of 
entity embeddings $\textbf{E}$ and relation embeddings $\textbf{R}$ are learned and $\textbf{R}$ will be used in axiom induction.

\vspace{-3mm}
\subsection{Axiom Induction}
Given relation embedding $\textbf{R}$, axiom induction aims to induce a set of axioms $\mathcal{A}$ and assign a confidence score to each axiom in $\mathcal{A}$. To achieve this, IterE firstly generates a pool of possible axioms $\mathcal{P}$ with an effective pruning strategy. Then it predicts a score of each axiom $a \in \mathcal{P}$  based on calculation with $\textbf{R}$.

\vspace{-3mm}
\subsubsection{Axiom pool generation}
Before calculating axiom scores with relation embedding $\textbf{R}$, relation variables $r, r_1, r_2$, and $r_3$ in Table \ref{linear_map_properties} should be replaced with concrete relations. 
Axiom pool generation generates a pool of possible axioms $\mathcal{P}$ by searching for possible axioms with the number of \emph{support} greater than $1$. 

One intuitive way of searching for possible axioms is generating axioms by replacing all relation variable in each type of axioms with each relation and then check the number of support for them, but this suffers from a huge search space. Another way is to generate axioms via random walk on knowledge graph while this method can't ensure coverage of axioms. Therefore axiom pool generation is not an easy task because it has to achieve a good balance between search space and coverage of highly possible axioms.

In this paper, we propose a pruning strategy combining traversing and random selection. 
There are two steps for generating possible axioms for each  relation $r \in \mathcal{R}$ in IterE. 
\vspace{-1mm}
\begin{itemize}[leftmargin=*, wide]
	\item \emph{step 1}: generate axioms or partial axioms: 
$\mathtt{ReflexiveOP}(r)$, 
$\mathtt{SymmetricOP}(r)$, 
$\mathtt{TransitiveOP}(r)$, 
$\mathtt{EquivalentOP}(r^\prime, r)$, 
$\mathtt{subOP}(r^\prime, r)$, 
$\mathtt{inverseOP}(r^\prime, r)$ and 
$\mathtt{subOP}(\mathtt{OPChain}(r^\prime, r^{\prime\prime}), r)$,
, in which $\{r^\prime, r^{\prime\prime}\}$ are relation variables to be replaced with relations in KG and $r$ participants in the head of their rule forms.

	\item \emph{step 2}: complete partial axioms via randomly selecting  $k$ triples $(e^\prime, r, e^{\prime\prime}) \in \mathcal{T}$ related with $r$.
	 Replacing $r ^ \prime$ or $r ^ {\prime\prime}$  in partial axioms with relations that directly link to $e^\prime$ or $e^{\prime\prime}$.
	 \item \emph{step 3}: search for support of each axiom, and add those axioms with number of support larger than $1$ into axiom pool $\mathcal{P}$.
	\end{itemize}
The key point of whole process is choosing $k$, which is not trivial because a large $k$ will lead to dramatic increase of search space while a small $k$ will decrease coverage of axioms immediately.
Thus a good choice of $k$ should achieve followings: 1) the probability for $\mathcal{P}$ covering all highly possible axioms higher than $t$, named \emph{including probability}. 2) $k$ is as small as possible. We defined the highly possible axioms as axioms with existing probability larger than $p$ in this paper and we name $p$ as \emph{minimum axiom probability}. 
We estimate the probability $p(r, a_x)$ that $r$ replaces the relation variable in \emph{head} of axiom $a_x$ as:
\vspace{-1mm}
\begin{equation}
\vspace{-1mm}
	 p(r, a_x) = \frac{n}{N}
\end{equation}
where $N$ is the number of triples $(e^\prime, r, e^{\prime\prime}) \in \mathcal{T}$ and $n$ is the number of support for axiom $a_x$ in $\mathcal{K}$. 
Thus choosing $k$ triples to satisfy two requirements mentioned above can be formulated as follows:
\vspace{-1mm}
\begin{equation}
\vspace{-1mm}
	\mathrm{min}\; k \quad
	s.t.\; 1- \frac{C_{(1-p)N}^k}{C_N^k} > t 
	\label{inequality}
\end{equation}
From above inequality in (\ref{inequality}), following inequality can be reached:
\vspace{-1mm}
\begin{equation}
\vspace{-1mm}
	 k > N - N (1-t)^{\frac{1}{pN}}
	 \label{inquality-result} 
\end{equation}
Thus with fixed $p,t$, the best choice of $k$ should be the upper bound $v_u$ of equation $f(N) = N - N (1-t)^{\frac{1}{pN}}$. 
Fortunately, with $N\in [0, 10^{15}]$, this equation is monotone increasing and has small upper bound $v_u$.
For example, when $ p = 0.5, t=0.95$, $v_u$ is $6$.  

As the input of axiom pool generation is a fixed $\mathcal{K}$ during learning, axiom pool $\mathcal{P}$ only need to be generated once. 

\vspace{-2mm}
\subsubsection{Axiom Predicting} 
Given current relation embedding $\textbf{R}$ and axiom pool $\mathcal{P}$, axiom predicting  
predicts a score $s_{a}$ for each axiom $a \in \mathcal{P}$ based on the  rule conclusion for each type of axioms (column 4 in Table \ref{linear_map_properties}), in the form that $\textbf{M}_1^a = \textbf{M}_2^a$, where $\textbf{M}_1^a$ and $\textbf{M}_2^a$ are matrices either from a single matrix or a product of two matrices. 
As rule conclusions are derived from ideal linear map assumption, $\textbf{M}_1^a$ and $\textbf{M}_1^a$ usually are not equal but similar during training.
Thus, instead of matching $\textbf{M}_1^a$ and $\textbf{M}_2^a$, 
we estimate the truth of axiom $a$ by similarity between $\textbf{M}_1^a$ and $\textbf{M}_2^a$ which is supposed to be related with Frobenius norm of their difference matrix:
\vspace{-1mm}
\begin{equation}
\vspace{-1mm}
 s_{a}(F) = \|\textbf{M}_1 ^a - \textbf{M}_2 ^a \|_F 
 \end{equation}
$s_{a}(F)$ is then normalized as follows because the value of $s_{a}(F)$ for different type of axioms vary dramatically:
\vspace{-1mm}
\begin{equation}
\vspace{-1mm}
s_a = \frac{s_{max}(t) - s_a(F)}{s_{max}(t) - s_{min}(t)}
\end{equation}
in which $t$ is the type of axiom that $a$ belongs to. 
$s_{max}(t)$ and $s_{min}(t)$ is the maximum and minimum Frobenius norm score among all type $t$ axioms in $\mathcal{P}$. $s_a \in [0,1]$ is the final score for axiom $a$ and the higher $s_a$ is the more confident that axiom $a$ is.

\vspace{-3mm}
\subsection{Axiom Injection}
\label{sec:axiom-injection}
Given knowledge graph $\mathcal{K}$ and possible axiom set $\mathcal{A}$, axiom injection utilizes axiom' deductive capability to infer a set of new triples $\mathcal{T}_{axiom}$ for sparse entities and predict their labels.  $\mathcal{T}_{axiom}$ will be injected into embedding learning to reduce the sparsity of entities.    

\vspace{-2mm}
\subsubsection{Sparse entities} 
We evaluate the sparsity of entities as follows:
\vspace{-2mm}
\begin{equation} 
\vspace{-1mm}
 sparsity(e) =  1- \frac{freq(e)-freq_{min}}{freq_{max}-freq_{min}} 
 \label{equ:entity-sparsity}
\end{equation}
where $freq(e)$ is the frequency of entity $e$ participating in a triple, as subject or object entity. 
$freq_{min}$ and $freq_{max}$ are the minimum and maximum frequency among all entities.
$sparsity(e) \in [0, 1]$.
$sparsity(e) = 1$ means $e$ is the most sparse entity and $sparsity(e) = 0$ means the most frequent one.
With $sparsity(e) > \theta _{sparsity}$, we regard entity $e$ as a sparse entity, where  $\theta _{sparsity}$ is a sparse threshold.
We use $\mathcal{E}_{sparse}$ to denote the set of sparse entity in $\mathcal{K}$.

During axiom injection, triples related with sparse entities are injected into the input of embedding learning. In other words, in new inferred triples $(s^a, r^a, o^a)$, either $s^a  \in \mathcal{E}_{sparse}$ or  $o^a \in \mathcal{E}_{sparse}$ or $\{ s^a, o^a\} \in \mathcal{E}_{sparse}$. 
Thus after inferring all possible new triples, we filter those unrelated to sparse entities.

\vspace{-2mm}
\subsubsection{Predicting new triples and their labels}
We utilize groundings to infer new triples, and the grounding for axioms considered in this paper can be generalized as the following form: 
\vspace{-1mm}
\begin{equation}
\vspace{-1mm}
	(s^a, r^a, o^a) \gets (s_1, r_1, o_1), (s_2, r_2, o_2),... , (s_n, r_n, o_n)
	\label{grounding}
\end{equation}
where the right side triples $(s_k, r_k, o_k) \in \mathcal{T}$ with $k \in [1, n]$ are generated from the body of axiom rule form and $(s^a, r^a, o^a) \notin \mathcal{T}$ is new inferred triples to be added into knowledge graph.

To predict label for $(s^a, r^a, o^a)$,  we first translate the grounding form in (\ref{grounding}) in propositional logical expression:
\vspace{-1mm}
\begin{equation}
\vspace{-1mm}
	(s_1, r_1, o_1) \land (s_2, r_2, o_2) \land...\land(s_n, r_n, o_n) \Rightarrow (s^a, r^a, o^a)
	\label{grounding2}
\end{equation}
then we model groundings through t-norm based fuzzy logics\cite{fuzzy-logic}.
It regards the truth value of a propositional logical expression as a composition of constituent triples' truth value, through specific logical connectives (e.g. $\land$ and $\Rightarrow$).
For example, the truth value of a propositional logical expression $(s_1, r_1, o_1) \Rightarrow (s_2, r_2, o_2)$ is determined by the true value of two component triples $(s_1, r_1, o_1)$ and $(s_2, r_2, o_2)$, via a composition defined by logical implication $\Rightarrow$.
We follow the definition of composition associated with logical conjunction ($\land$), disconjunction ($\lor$) and negation ($\lnot$) as \cite{RUGE:conf/aaai/GuoWWWG18}: 
\vspace{-1mm}
\begin{align}
\vspace{-1mm}
\pi (a \land b) &= \pi(a) \cdot \pi(b) \label{t-norm-equiation1}\\
\pi(a \lor b) &= \pi(a) + \pi(b) - \pi(a)\cdot\pi(b) \\
\pi(\lnot a) &= 1-\pi(a)\\
\pi(a \Rightarrow b) &= \pi(\lnot a \lor b)   \label{t-norm-equiation4}
\end{align}
in which $a, b$ are logical expressions and $\pi(x)$ is the truth value of $x$.
Given these compositions, the truth value of any propositional logical expression can be calculated recursively through Equation (\ref{t-norm-equiation1})-(\ref{t-norm-equiation4}). For example, applying to propositional logical expression of grounding $g:\; (s_1, r_1, o_1) \land (s_2, r_2, o_2)  \Rightarrow (s^a, r^a, o^a)$:
$$\pi(g) 
= 1- \pi(s_1, r_1, o_1)\pi(s_2, r_2, o_2) + \pi(s_1, r_1, o_1)\pi(s_2, r_2, o_2)\pi(s^a, r^a, o^a) $$ 

To predict the truth value $\pi(s^a, r^a, o^a)$ for triples $(s^a, r^a, o^a)$, inferred by axiom $a$ according to grounding $g_a$, based on t-norm fuzzy logics, $\pi(s^a, r^a, o^a)$ can be calculated according to $\pi(g)$ and $\pi(s_1, r_1, o_1)$, $\pi(s_2, r_2, o_2),... , \pi(s_n, r_n, o_n)$  $ \in g_a$.
In former example,
$$ \pi(s^a, r^a, o^a) = \frac{\pi(g) - 1 + \pi(s_1, r_1, o_1)\pi(s_2, r_2, o_2)}{\pi(s_1, r_1, o_1)\pi(s_2, r_2, o_2)}  $$

For the truth value of axiom $a$'s grounding $g$, we evaluate it as the score of $a$ generated by axiom induction, namely $\pi(g_a) = s_a$. 
For the truth value of triples existing in knowledge graph, we evaluate them with their training labels, thus $\pi(s, r, o) = 1 \;\mathrm{for}\; (s, r, o) \in \mathcal{T}$, because existing triples are absolutely true. 
With these two types of truth value assignment, 
we can easily get the following result for triples $(s^a, r^a, o^a)$ inferred by any type of axioms $a$:
\vspace{-1mm}
\begin{equation}
\vspace{-1mm}
\pi(s^a, r^a, o^a) = s_a	
\end{equation}

After axiom injection, a set of new triples $\mathcal{T}_{axiom} = \{(s^a, r^a, o^a)|$ 
$s^a \in \mathcal{E}_{sparse}\  \mathrm{or}\  o^a \in \mathcal{E}_{sparse} \}$ are inferred via quality axioms and each new triple $(s^a, r^a, o^a)$ is labeled with $l_{s r o} = s_a$. Thus the input of embedding learning was updated, $\mathcal{I} = \{((s, r, o), l_{sro}) | (s, r, o ) \in {\mathcal{T} \land \mathcal{T}_{axiom} } \}$. Then the process goes back to embedding learning.

\vspace{-2mm}
\section{Experiment}
\label{sec:experiment}
During experiments, we want to explore following questions:
1) whether axioms really help sparse entity embedding learning? To do this, we evaluate the quality of embeddings on link prediction task which is widely applied in previous knowledge graph embedding works;
2) whether embeddings really help rule learning overcome the huge search space and improve
the quality of learned rules? To do this,  we evaluate the efficiency of axiom learning with learning time and the quality with number and percentage of high quality axioms the method generate; 
3) how does the iterative manner affect embedding learning and rule learning process?  To this end, we show  the changes of link prediction results, rule qualities and the number of triples injected along with different number of iterations.
\vspace{-3mm}
\subsection{Dataset}
Four datasets are used in experiment, including  WN18-sparse, WN18RR-sparse, FB15k-sparse, and FB15k-237-sparse. They are generated from WN18\cite{TransE:conf/nips/BordesUGWY13}, WN18RR\cite{ConvE:conf/aaai/2018}, FB15k\cite{TransE:conf/nips/BordesUGWY13} and FB15-237\cite{toutanova2015observed}, four datasets that are commonly used in previous knowledge graph embedding work.
WN18 and WN18RR are subsets of WordNet, a lexical knowledge graph for English. FB15k and FB15k-237 are subsets of a large collaborative knowledge base Freebase. The statistics of datasets are listed in Table \ref{tab:datasets}.

We generate the sparse version of these datasets via changing the valid and test datasets which will be used for link prediction evaluation. In link prediction experiments, we want to explore whether axioms really contribute to sparse entity embeddings. 
Therefore, we only keep triples in test and valid dataset which involve at least one sparsity entity. In other words, for a test or valid triple $(s, r, o)$, if either $s$ or $o$ is or both are sparse entity (Equation (\ref{equ:entity-sparsity})), the triple will be kept in sparse dataset, otherwise, it will be filtered.
\begin{table}[!htbp]
\centering
\footnotesize
\caption{\small{Statistics of datasets.} }
\vspace{-4mm}
  \begin{tabular}{ c |c  c  c  c c }
    \toprule
      \textbf{Dataset} & \textbf{\#E} & \textbf{\#R} & \textbf{\#Train} 
      & \textbf{\#Valid} 
      & \textbf{\#Test}\\
      \midrule
      WN18-sparse   & 40,943 & 18 & 141,442 
      & 3624(72.48\%)  
      & 3590(71.8\%) \\
      WN18RR-sparse & 40943 & 11 & 86835 
      & 1609(53.03) 
      & 1661(52.9\%)\\
      FB15k-sparse  & 14,951 & 1,345 & 483,142 
      & 18544(37.08\%) 
      & 22013(37.26\%) \\
      FB15k-237-sparse  & 14541 & 237 & 272115 
      &10671(60.8\%) 
      &12454(60.85\%)\\
       
    \bottomrule
  \end{tabular}
  
\label{tab:datasets}
\vspace{-5mm}
  \end{table}
 
When deciding sparse entity hyperparameter $\theta$, intuitively a principle is considered: the percentage of left triples in valid and test set should not be larger than $80\%$ or less than $20\%$ of original valid and test set for all datasets. 
Thus we choose $\theta = 0.995$ and regard entities $e$ with $sparsity(e)>0.995$ as sparse entity in this paper. Table \ref{tab:datasets} also shows the percent of triples left.

\vspace{-3mm}
\subsection{Training Details}
For embedding learning, 
the number of negative samples is set to $6$ and the number of scalars on the diagonal of each relation matrix is  set to $\frac{d}{2}$ where $d$ is embedding dimension. We initialize embeddings with uniform distribution $\textbf{U}(-0.1, 0.1)$.

For axiom induction,  we set the minimum axiom probability $p = 0.5$ and the including probability $t=0.95$ for axiom pool generation. Based on these settings, the number of related triples selected for each relation is set as $k=6$ according to Equation(~\ref{inquality-result}).
Details of axiom pools generated for all datasets are shown in Table \ref{tab:axiom-pools}, where we can see that the number of possible axioms in FB15k and FB15k-237 are much larger than WN18 and WN18RR because the number of axioms is highly related to the diversity of relations.
\begin{table}[!htbp]
\centering
\footnotesize
\vspace{-3mm}
\caption{\small{Axiom pools details.}}
\vspace{-4mm}
\begin{tabular}{c| c|c| c|c| c|c| c}
\toprule
  		& \textbf{ref.} & \textbf{sym.} & \textbf{tra.} & \textbf{equ.} & \textbf{inv.} & \textbf{sub.} & \textbf{sub.(Cha.)} \\
 \midrule
 WN18-sparse
 & 0 		& 3 		&	1 		 & 2		  & 19		&	2		  &		72			\\
 
 WN18RR-sparse 
 & 0 		& 3 		&	1 		  & 1		&	3		  &	1		&22	\\
 
 FB15k-sparse
 & 41	& 94		& 52		 &1872		  & 2743 	&  1872		  & 59709 \\
 FB15k-237-sparse 
 & 5 & 30 & 27 &197  & 192 & 197 & 5017 \\
 \bottomrule
\end{tabular}

\label{tab:axiom-pools}
\vspace{-4mm}
\end{table}



For axiom injection, considering that axioms with high scores are more reliable and less possible to introduce noise, 
we set a threshold $\theta$ for each dataset and regard axioms with scores $s_{axiom} > \theta$ as high quality axioms. We also set a maximum inferred triples $m$ for axioms in each dataset, if one axiom will infer a lot of triples we ignore it because a huge number of triples inferred by one axiom will change the data distribution significantly and make the embedding training unstable.

During one iteration learning, we first train embedding learning for 10 epochs, and then conduct axiom induction and injection once. The maximum training iteration is set to $10$ for WN18RR-sparse and FB15k-237-sparse and $50$ for WN18-sparse and FB15k-sparse. 
We use Adam algorithm~\cite{Adam:journals/corr/KingmaB14} during optimization with learning rate $lr = 0.001$.
We apply grid search for the best hyperparameters based on the filter MRR on the validation dataset, with combinations from embedding dimension $d \in \{100, 200\}$ and $l_1$ regularizer weigh $\lambda \in \{10^{-4},10^{-5}, 10^{-6}\}$.  

The final parameters are $d=200, \lambda=10^{-5}, \theta =0.95,  m = 20000$ for WN18-sparse, $d=200, \lambda=10^{-5}, \theta =0.9, m = 10000$ for WN18RR-sparse, $d=200, \lambda=10^{-4}, \theta =0.95 , m = 100$ are for FB15k-sparse, and $d=200, \lambda=10^{-5},\theta =0.9, m = 1000$ are for FB15k-237-sparse.  
\vspace{-3mm}
\subsection{Embedding Evaluation}
\begin{table*}[htb]
\small
\centering
\caption{
\small{Link prediction results with MRR and Hit@n on WN18RR-sparse and FB15k-237-sparse. Underlined scores are the better ones between ANALOGY and IterE(ANALOGY). Boldface scores are the best results among all methods.} 
}
\vspace{-4mm}
  \begin{tabular}{ l |c c  c c c | c c  c c c }
    \toprule
    \multirow{3}{*}{} & \multicolumn{5}{|c|}{\textbf{WN18-sparse}} &  \multicolumn{5}{|c}{\textbf{FB15k-sparse}} \\
    \cline{2-11}	& MRR		&MRR		&Hit@1		&Hit@3	&Hit10	&MRR		&MRR		&Hit@1		&Hit@3	&Hit10	\\
    & (filter)	&(raw)		&(filter)		&(filter)	&(filter)		&(filter) &(raw)	&(filter)	&(filter)	&(filter)\\
    \hline
    TransE\cite{TransE:conf/nips/BordesUGWY13}	 		&41.8 &33.5 &10.2 &71.1 &84.7
    				&39.8 &25.5 &25.8 &48.6 &64.5	\\
    DistMult\cite{DistMult:conf/iclr/2015} 	&73.8 &55.8 &59.3 &87.5 &93.1
    				&60.0 &32.4 & \textbf{61.8} &65.1 &75.9		\\
    ComplEx\cite{ComplEx:conf/icml/TrouillonWRGB16}	&91.1 &67.7 &89.0 &93.3 &94.4
    				&61.6 &32.7 &54.0 &65.7 & 76.1	\\

    \midrule
    ANALOGY\cite{ANALOGY:conf/icml/LiuWY17} &\underline{\textbf{91.3}} &\underline{67.5} &\underline{89.0} &\underline{93.4} & 94.4
    				&\underline{62.0} &33.1&\underline{54.3} &66.1 &76.3	\\
    
    				
   	 \textbf{IterE}(ANALOGY)  &90.1 &\underline{67.5} &87.0 &93.1 &\underline{\textbf{94.8}}
    				&61.3 &\underline{35.9} &52.9 &\underline{66.2} &\underline{76.7} \\
    				
    \textbf{IterE}(ANALOGY) + \textbf{axioms} &\textbf{91.3} &\textbf{78.9} &\textbf{89.1} &\textbf{93.5} & \textbf{94.8}
    				&\textbf{62.8} &\textbf{38.8} &55.1 &\textbf{67.3} &\textbf{77.1}	\\
    \toprule

    \multirow{3}{*}{} & \multicolumn{5}{|c|}{\textbf{WN18RR-sparse}} &  \multicolumn{5}{|c}{\textbf{FB15k-237-sparse}} \\
    \cline{2-11}	& MRR		&MRR		&Hit@1		&Hit@3	&Hit10	&MRR		&MRR		&Hit@1		&Hit@3	&Hit10	\\
    & (filter)	&(raw)		&(filter)		&(filter)	&(filter)		&(filter) &(raw)	&(filter)	&(filter)	&(filter)\\
    \hline
    TransE\cite{TransE:conf/nips/BordesUGWY13}	 		&14.6 &12.4 &3.4 &24.7 &28.8
    				&23.8 &15.6 &16.4 &26.1 &38.5	\\
    DistMult\cite{DistMult:conf/iclr/2015} 		&25.5 & 20.8 &23.8 & 26.0 & 22.5
    				&20.4 &12.9 &12.8 &22.6 &36.2 \\
    ComplEx\cite{ComplEx:conf/icml/TrouillonWRGB16}	&25.9		&21.4		&24.6		&26.2 	&28.6	
    		&19.7 		&13.3		&12.0		&21.7	&35.4	\\

    \midrule
    ANALOGY\cite{ANALOGY:conf/icml/LiuWY17} &19.8		&13.3		&24.6		&27.5		&28.7	
    		&19.8 		&13.9		&12.3 		&21.4		&34.9\\
    
    \textbf{IterE}(ANALOGY)   &\underline{27.2}  &\underline{22.7}  &\underline{25.0} &\underline{\textbf{28.1}} &\underline{\textbf{31.4}} 
    				  &\underline{20.7} &\underline{14.0} &\underline{13.1} &\underline{22.8}  &\underline{36.2} \\
    \textbf{IterE}(ANALOGY) + \textbf{axioms} &\textbf{27.4} &\textbf{25.7} &\textbf{25.4} &\textbf{28.1} &\textbf{31.4} 
    						 &\textbf{24.7} &\textbf{18.6} &\textbf{17.9} &\textbf{26.2} &\textbf{39.2} \\
    \bottomrule

\end{tabular}

\vspace{-3mm}
\label{Link Prediction}
\end{table*}
\begin{table*}[!htbp]
\centering
\small
\caption{Rule evaluation results.\label{tab:rule-evaluation-results}}
\vspace{-4mm}
  \begin{tabular}{ c |c c  c | c  c c |c c  c | c  c c}
    \toprule
      \multirow{2}{*}{}  
      	& \multicolumn{3}{c|}{\textbf{WN18-sparse}} 
      	& \multicolumn{3}{c|}{\textbf{WN18RR-sparse}}
      	& \multicolumn{3}{c|}{\textbf{FB15k-sparse}}  
      	&\multicolumn{3}{c}{\textbf{FB15k-237-sparse}} \\
      \cline{2-13}
      & time  &HQr & \%  & time &HQr & \% & time  &HQr & \%  & time &HQr & \%\\
      \hline
      \textbf{AMIE+}   &4.98s &16 &11.4\% 
       
      		&3s &2  &  5.71\% 
      		 &428s &1820 &4.4\%  
      		&66s &470  &1.9\% \\
      \textbf{IterE}   &\textbf{1.63}s &\textbf{20} &\textbf{20.2}\% 
      &\textbf{0.75}s &\textbf{6} &\textbf{19.3}\% 
      &\textbf{26.49}s &\textbf{11375} & \textbf{17.6}\% 
       &\textbf{4.72}s  &\textbf{653}  &\textbf{11.8}\%   \\
    \bottomrule
  \end{tabular}
  \vspace{-4mm}
 \end{table*}
  

We evaluate the quality of embeddings on link prediction tasks.
Link prediction aims to predict the missing entity when given the other entity and relation in a triple, including subject entity prediction $(?, r, o)$ and object entity prediction $(s, r, ?)$. 

\subsubsection{Evaluation metrics} Given subject entity prediction task $(?, r, o)$ with right answer $s$, we first fit subject entity position with each entity $e \in \mathcal{E}$ and thus get a set of triples $\mathcal{T}_{subjectprediction} = \{(e, r, o) | e \in \mathcal{E} \}$. Then we calculate the score for each triple in $\mathcal{T}_{subjectprediction}$ according to Equation (\ref{score-function}) and rank their scores in descending order. Thus the entity $e$ in $(e, r, o)$ with a higher rank is a more possible prediction. 
To evaluate how good the prediction is, we use the rank of $(s, r, o)$ among all triples in $\mathcal{T}_{subjectprediction}$ as subject entity prediction evaluation result for $(s, r, o)$, namely subject entity prediction rank $rank_s(?, r,o)$.
The object entity prediction task is done in the same way and will get objection entity rank $rank_o(s, r, ?)$. 
The final prediction rank $rank(s, r, o)$ for $(s, r, o)$ is the average of subject and object prediction rank:
\vspace{-2mm}
$$ rank(s, r, o) = \frac{1}{2}(rank_s(?, r,o) + rank_o(s, r, ?))$$
Aggregating prediction rank for all test triples, we applied mean reciprocal predicting rank  \emph{(MRR)} and the percentage of predicting ranks within $n$ \emph{(Hit@n)} to evaluate the whole prediction result. These two metrics are widely used in previous work \cite{ComplEx:conf/icml/TrouillonWRGB16}\cite{ANALOGY:conf/icml/LiuWY17}.
Generally, a higher \emph{MRR} or \emph{Hit@n} indicates a better prediction result.

We also apply $filter$ and $raw$ setting. As mentioned in \cite{TransH:conf/aaai/WangZFC14}, when fitting subject or object entity position with other entities, we may generate triples existing in knowledge graph which are known to be true. 
It is not wrong if these triples are ranked higher than current test triple. With $filter$ setting, we filter the generated triples that exist in both train/valid/test dataset before ranking but not current test triple. $raw$ means the setting of NO filtering.

\vspace{-1mm}
\subsubsection{Baselines}
For baselines, 
one method from translation-based assumption  TransE
\footnote{the code for TransE is from https://github.com/thunlp/OpenKE. We train all dataset with learning rate $0.01$, margin $1.0$ and dimension $100$ for maximum 3500 iterations.} 
and three methods based on linear map assumption, DistMult, ComplEx, and ANALOGY
\footnote{The code for DistMult, ComplEx and ANALOGY is from https://github.com/quark0/ANALOGY. 
We train them with the same parameter setting as \cite{ANALOGY:conf/icml/LiuWY17} for maximum 500 iterations.
For WN18-sparse, the parameters  are dimension $d = 200$, regularizer weigh $\lambda = 10^{-1}$, negative weight $n = 6$ for DistMult and $d = 200, \lambda = 10^{-2}, n  =6$ for both ComplEx and ANALOGY.
For WN18RR-sparse, the parameters are $d = 200, \lambda=1-^{-1}, n =3$ for DistMult, Complex and ANALOGY.
For FB15k-sparse, the parameters are $d = 200, \lambda=10^{-2}, n = 6$ for DistMult, Complex and ANALOGY.
For FB15k-237-sparse, the parameters are dimension $d = 100,\lambda = 0.1, n = 6$ for ANALOGY and Complex and $d=200, \lambda=10^{-1}, n=3$ for DistMult. 
} are considered.

\vspace{-1mm}
\subsubsection{Results and analysis}
To show how axiom helps sparse entity embeddings, we adopt two strategies: 
(1) Firstly, we evaluate how axioms directly improve entity embedding quality and compare link prediction results from our method, denoted as \textbf{IterE} in Table \ref{Link Prediction}, with other baselines directly. (2) Secondly, we evaluate how axioms can help improve sparse entity link prediction utilizing its deductive ability. Thus we compare prediction results with embeddings and axioms, denoted as \textbf{IterE+axioms} in Table \ref{Link Prediction} with IterE and other methods. In IterE+axioms, if the test triple $(s, r, o)$ are inferred by axioms during axiom injection, which means $(s, r, o) \in \mathcal{T}_{axiom}$ , we regard it as correct and mark its prediction rank as $1$. 

The link prediction results are shown in Table~\ref{Link Prediction}. We analyze the results as follows: Firstly, the link prediction results of IterE competitive to ANALOGY, which means most of the triples injected into embedding learning are not noise, indicating that learning axioms from embeddings works well. 
Secondly, IterE outperforms baselines on WN18RR-sparse and FB15k-237-sparse, while are slightly improved on WN18-sparse and FB15k-sparse. 
These indicate that IterE helps sparse entities much more in WN18RR-sparse and FB15k-237-sparse than in WN18RR-sparse and FB15k-sparse.
This is quite reasonable in our opinion, because among these four datasets, WN18RR-sparse and FB15k-237-sparse are more challenging and more sparse. They are created from WN18-sparse and FB15k-sparse via removing one relation of all inverse relation pair in the dataset and also their related triples because inverse relation pair is a significant pattern in these two datasets as first noted in \cite{toutanova2015observed}.
Thirdly, the results of IterE+axioms are improved compared with IterE in all datasets, especially on the most complex and sparse dataset FB15k-237-sparse. It indicates that the deductive capability of axioms can help the prediction of sparse entities further.

From the evaluation of link prediction, we can conclude that: 
(1) by injecting new triples for sparse entities, axioms help improve the quality of sparse entity embeddings and are more helpful in sparse KGs.
(2) Combining axioms and embeddings together to predict missing links works better than using embeddings only. Both the deductive capability of axioms and the inductive capability of embeddings contribute to prediction and complement each other.

\vspace{-3mm} 
\subsection{Rule Evaluation}

We evaluate the learned rules/axioms from two perspectives: efficiency and quality. We compare our method with AMIE+\cite{AMIE+:journals/vldb/GalarragaTHS15}\footnote{we run AMIE+ code from  https://www.mpi-inf.mpg.de/departments/databases-and-information-systems/research/yago-naga/amie/} which is an improved rule mining system of widely used AMIE\cite{AMIE:conf/www/GalarragaTHS13}.

\vspace{-2mm}
\subsubsection{Evaluation metrics}
The efficiency of rule learning is evaluated by learning time.
The quality of rule learning is evaluated with the number of high quality rule (HQr) and their percentage. 
The quality of rules are evaluated by head coverage(HC) which is commonly used in pervious work, such as \cite{AMIE:conf/www/GalarragaTHS13} and \cite{RLvLR:conf/ijcai/OmranWW18}. 
Head coverage for rule $rul$ is defined as follows:
\vspace{-2mm}
$$HC(rul) =  \frac{\# (e, e^\prime): support(rul)\land head(rul)(e, e^\prime) }{\# (e, e^\prime):head(rul)(e, e^\prime)}$$
in which $head(rul)(e,e^\prime) = \{ (e, r, e^{\prime}) \in \mathcal{T}\}$ if the head atom of $rul$ is $(\mathrm{X}, r, \mathrm{Y})$. 
And $support(rul)$ is the supports for $rul$. 
We regard high quality rules as the rules with $HC > 0.7$ during test.

\vspace{-2mm}
\subsubsection{Results and analysis}
Rule evaluation results are shown in Table \ref{tab:rule-evaluation-results}. According to the time used to generate rules among all datasets, we can see that IterE learns rules more efficiently. 
For example, with FB15k-sparse and FB15k-237-sparse datasets, IterE costs 10 times less than AMIE+. 
This shows our pruning strategy works well.
Among 4.72 seconds IterE cost for FB15k-237-sparse dataset, there is 4.55 seconds used for axiom pool generation and 0.17 seconds for axiom score calculation, namely axiom score calculation only cost $3.6\%$ of the time. 
This means calculating axiom scores via embeddings is super efficient.
We didn't include the time of embedding learning during this evaluation, as embedding learning is not devised mainly for rule learning, but for link prediction. 
The number of high quality rules shows that IterE generates more high quality rules than AMIE+ for each dataset and also achieves a higher percentage. 
This indicates that our axiom pool generation  can filter meaningless axioms(rules) and achieve a good balance between small search space and coverage of highly possible axioms. 

\begin{figure}[!hptb]
\centering
\subfigure[]{
\includegraphics[scale=0.35]{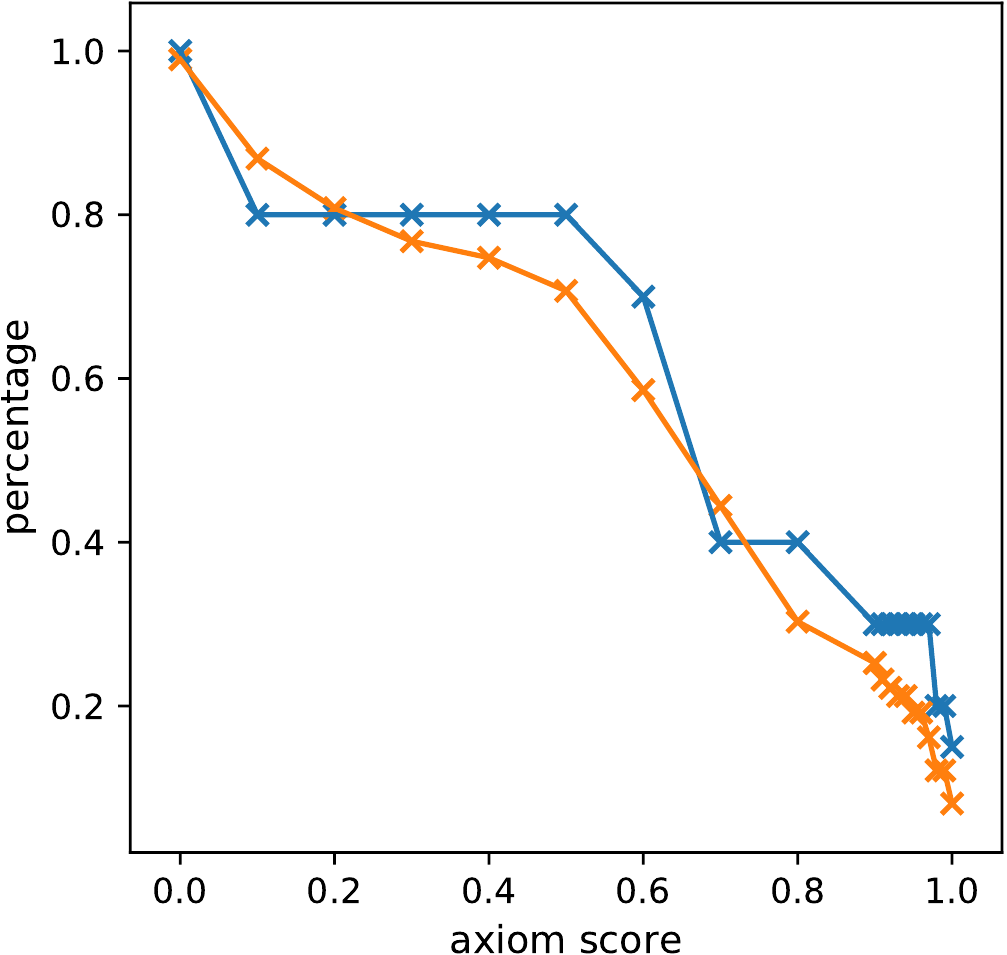}
\label{WN18}
\hspace{-1mm}
}
\subfigure[]{
\includegraphics[scale=0.35]{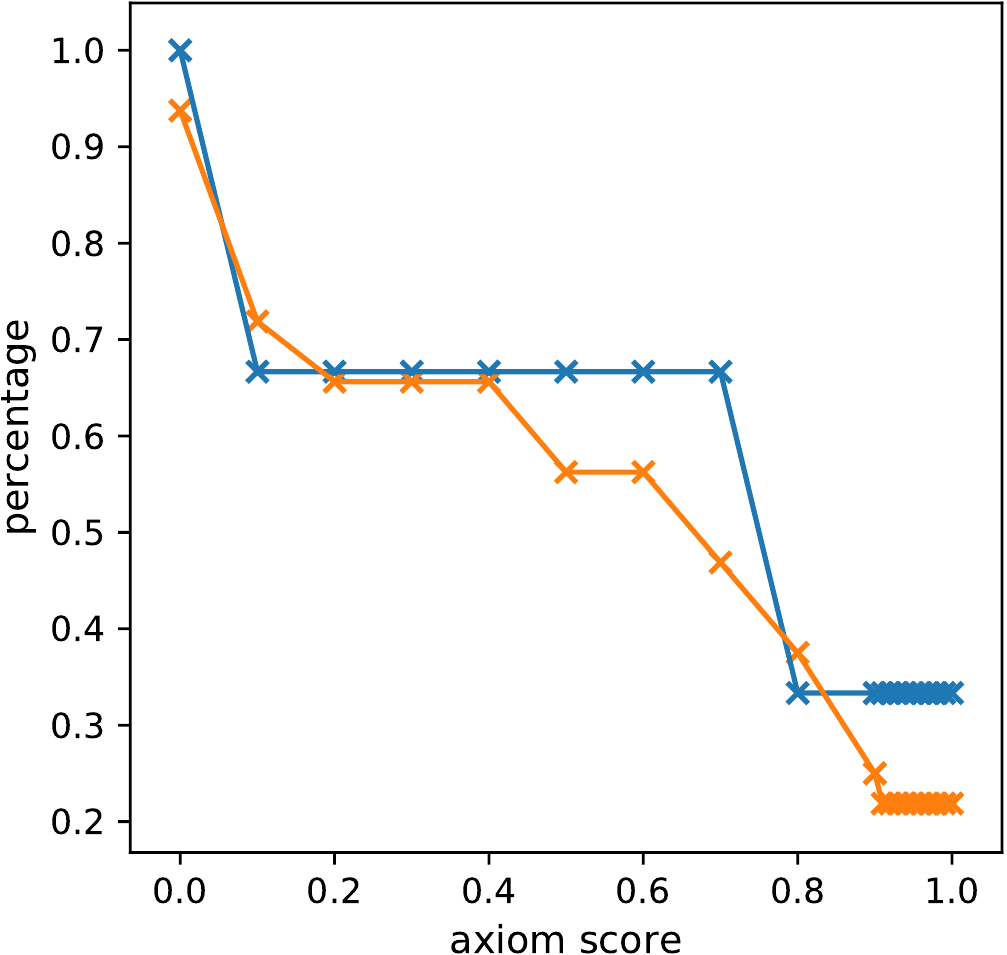}
\label{WN18RR}
}
\subfigure[]{
\includegraphics[scale=0.35]{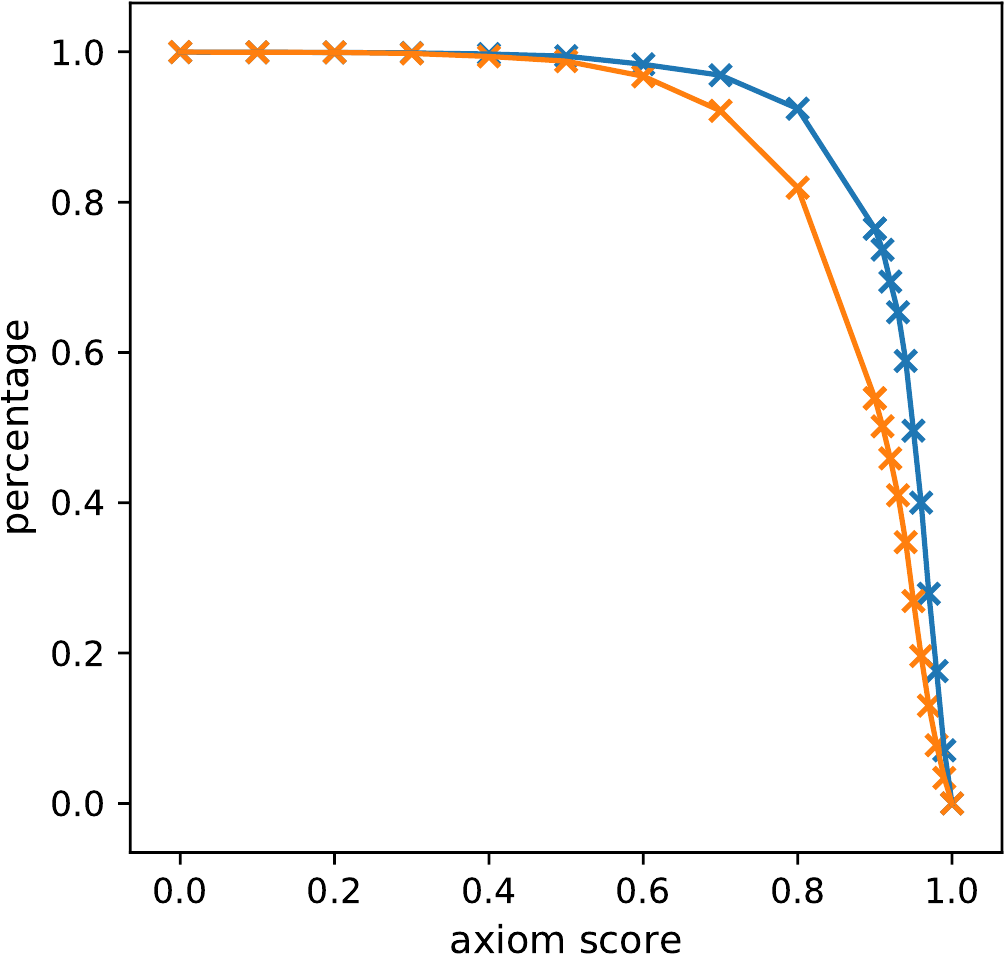}
\label{FB15k}
\hspace{-1mm}
}
\subfigure[]{
\includegraphics[scale=0.35]{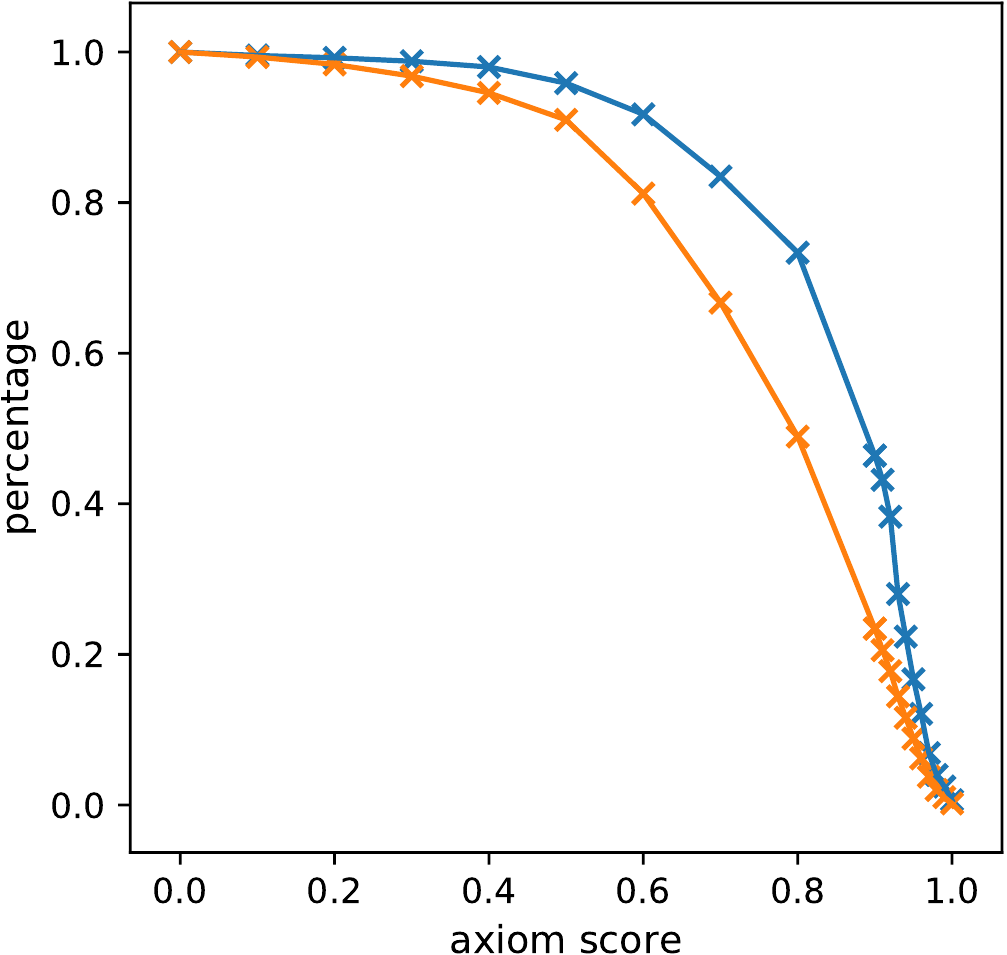}
\label{FB15k-237}
}
\vspace{-6mm}
\caption{\small{The coverage of high quality axioms (blue) and the axiom percentage (orange)
with different axiom score thresholds. \ref{WN18} is from WN18-sparse. \ref{WN18RR} is from WN18RR-sparse. \ref{FB15k} is from FB15k-sparse. \ref{FB15k-237} is from FB15k-237-sparse.} }
\label{fig:axiom-score-percentage} 
\vspace{-4mm}
\end{figure}

Further more, Figure~\ref{fig:axiom-score-percentage} shows the changes of high quality axioms coverage and axiom percentage with different axiom score thresholds from IterE.
For example, in \ref{FB15k}, with axiom threshold $0.9$, which means we select axioms with $s_a>0.9$ from IterE for axiom injection, there are $23.4\%$ axioms selected among all axioms and $46.4\%$ high quality axioms included. And in \ref{FB15k-237}, with axiom threshold $0.9$, there are $53.9\%$ axioms selected and $76.5\%$ high quality axioms included.  
It illustrates that axiom scores calculated from embeddings are 
reliable because it is consistent with rule evaluation results. 
The results in \ref{WN18} and \ref{WN18RR} for WN18-sparse and WN18RR-sparse are not obvious because a few relations are contained in these two datasets and the number of learned rules is limited. 

From rule evaluation results, we can conclude that (1) embeddings together with axiom pool generation help rule learning overcome large search space problem and improve rule learning efficiency, and (2) they also improve rule learning qualities and rules' reliable scores generated based on calculation with embeddings.

\vspace{-2mm}
\subsection{Iterative learning}
\begin{figure}[!hptb]
\centering
\vspace{-4mm}
\subfigure[]{
\includegraphics[scale=0.35]{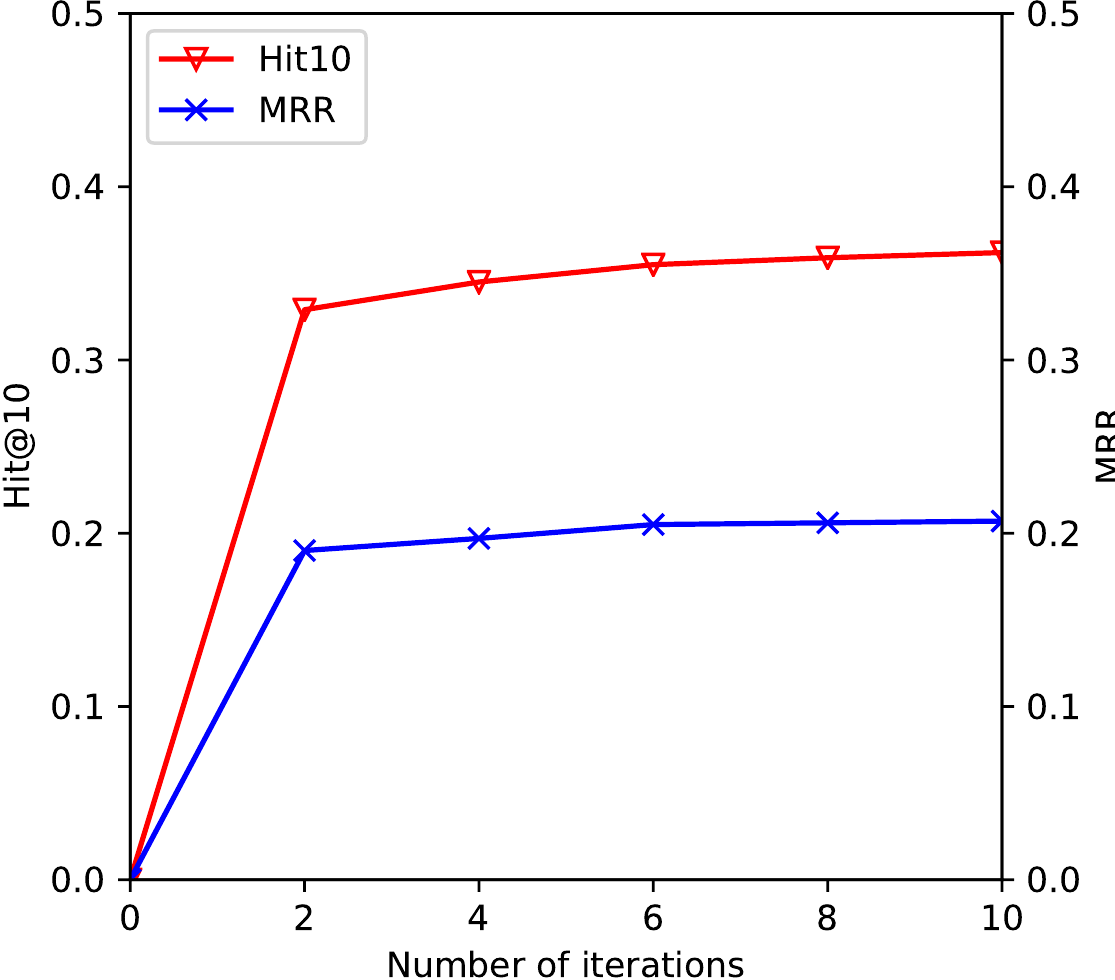}
\label{iter-link-predidction}

}
\subfigure[]{
\includegraphics[scale=0.35]{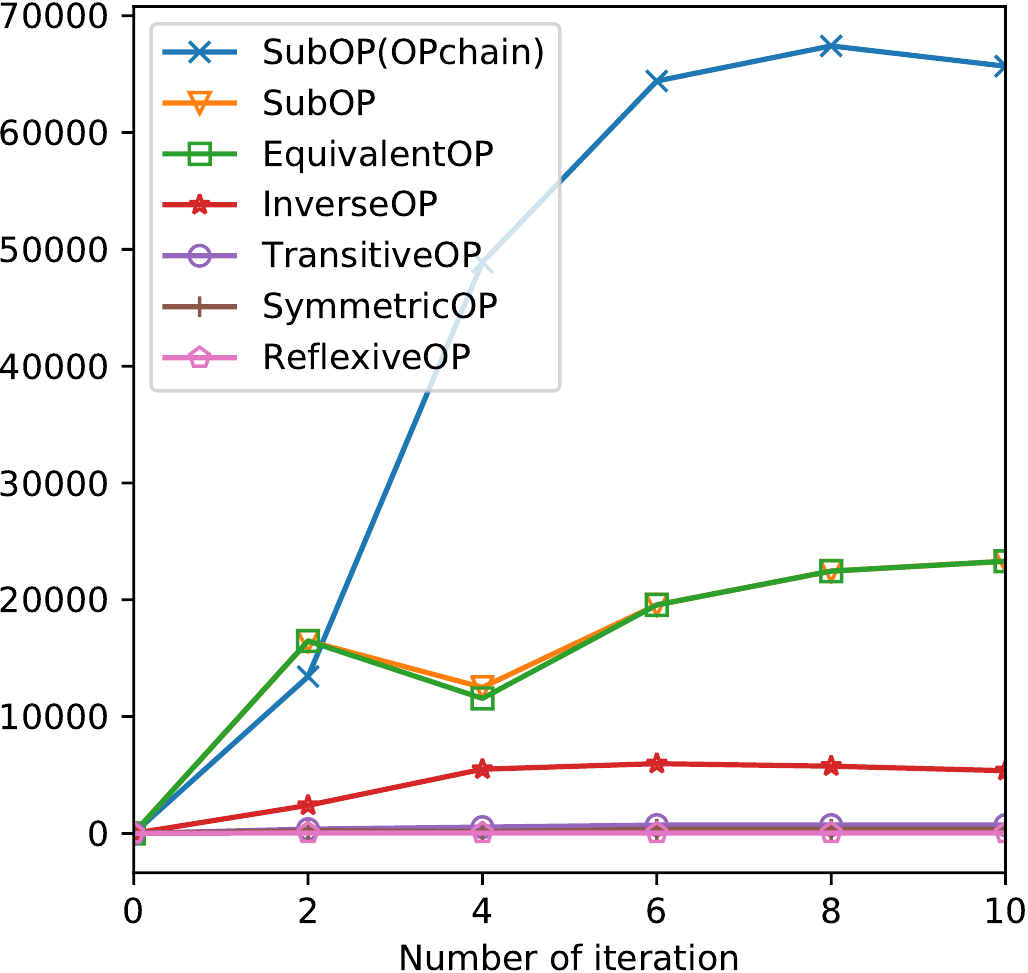}
\label{iter-injected-triples}
}
\vspace{-5mm}
\caption{\small{\ref{iter-link-predidction} shows link prediction results, MRR and Hit@10, in different iterations. \ref{iter-injected-triples} shows the number of triples injected into embedding learning in different iterations.} }
\vspace{-4mm}
\label{fig:iter-evaluation} 
\end{figure}

\vspace{-1mm}
To explore how iterative training improves embedding and rule learning during training, we show link prediction results on FB15k-237-sparse and number of injected triples at different iterations in Figure \ref{fig:iter-evaluation}. 

\begin{table*}
\small
\centering
\caption{
	\small{Examples from FB15k-237-sparse about the link prediction changes of test triples involving sparse entity. Following details are shown: 
the sparsity of subject and object entity in test triple, number within the square bracket beside the entity; 
the axiom that infers the test triple;
the change of link prediction rank from ANALOGY to IterE;
the change of link prediction rank from IterE to IterE+axiom.}
	}
\vspace{-3mm}
	\begin{tabular}{ l | c| c | c |c }
	\toprule
	\textbf{triple} & \multicolumn{4}{c}{(\texttt{Groundhog\_Day}[$0.996$]),  $/film/currency$,  \texttt{United\_States\_Dollar}[$0.615$])} \\
	\hline
	\textbf{predicted by axiom} &  \multicolumn{4}{c}{(\texttt{X}, $/film/currency$, \texttt{Z}) $\gets$ (\texttt{X}, $film\_release\_region$, \texttt{Y}), (\texttt{Y}, $/location/currency$, \texttt{Z}) } \\
	\hline
	\textbf{rank change}(ANALOGY $\to$ IterE) & \textbf{subject prediction rank:} & 1409 $\to$ 356 (+1053)  &\textbf{object prediction rank:} & 1 $\to $ 1(+0) \\
	\textbf{rank change}(IterE $\to$ IterE+axiom) & \textbf{subject prediction rank:} & 356 $\to$ 1 (+355)  &\textbf{object prediction rank:} & 1 $\to $ 1(+0) \\
	\toprule
	
	\textbf{triple} & \multicolumn{4}{c}{(\texttt{USA}[$0.0$]),  $/second\_level\_divisions$,  \texttt{Champaign\_County}[$0.999$])} \\
	\hline
	\textbf{predicted by axiom} &  \multicolumn{4}{c}{(\texttt{X}, $/second\_level\_divisions$, \texttt{Z}) $\gets$ (\texttt{Y}, $/bibs\_location$, \texttt{X}), (\texttt{Z}, $/county\_seat$, \texttt{Y})} \\
	\hline
	\textbf{rank change}(ANALOGY $\to$ IterE) & \textbf{subject prediction rank:} & 1 $\to$ 1 (+0)  &\textbf{object prediction rank:} & 416 $\to $ 214(+202) \\
	\textbf{rank change}(IterE $\to$ IterE+axiom) & \textbf{subject prediction rank:} & 1 $\to$ 1 (+0)  &\textbf{object prediction rank:} & 214 $\to $ 1(+213) \\
	
	\toprule
	
	\textbf{triple} & \multicolumn{4}{c}{(\texttt{Jenny\_McCarthy}[$0.997$]),  $/type\_of\_union$,  \texttt{Marriage}[$0.607$])} \\
	\hline
	\textbf{predicted by axiom} &  \multicolumn{4}{c}{(\texttt{X}, $/type\_of\_union$, \texttt{Z}) $\gets$ (\texttt{X}, $spouse$, \texttt{Y}), (\texttt{Y}, $/type\_of\_union$, \texttt{Z}) } \\
	\hline 
	\textbf{rank change}(ANALOGY $\to$ IterE) & \textbf{subject prediction rank:} & 4199 $\to$ 33 (+4166)  &\textbf{object prediction rank:} & 2 $\to $ 2(+0) \\
	\textbf{rank change}(IterE $\to$ IterE+axiom) & \textbf{subject prediction rank:} & 33 $\to$ 1 (+32)  &\textbf{object prediction rank:} & 2 $\to $ 1(+1) \\
	\bottomrule
	\end{tabular}
	
	\label{sparse-entity-prediction-examples}
	\vspace{-3mm}
\end{table*}

Figure~\ref{fig:iter-evaluation} shows that the prediction results including Hit@10 and MRR become better as the training iteration increases and generally the number of injected triples increases during training and finally gets stable. 
The number of injected triples for $\mathtt{subOP}$ and $\mathtt{equivalentOP}$ in iteration $4$ is less than iteration $3$ which means axioms with high scores learned from previous iteration might get a lower score in the next iteration. This indicates that embedding learning effects rule learning.
Thus from Figure~\ref{fig:iter-evaluation}, 
we can conclude that:
(1) Iterative learning benefits embedding learning as the quality of embeddings gets better during training.
(2) Iteratively learning benefits axiom learning because more axioms are learned and more triples are injected during training. 
(3) Axioms and embeddings influence and constrain each other during training.

\vspace{-3mm}
\subsection{Case study}
\vspace{-1mm}
In Table~\ref{sparse-entity-prediction-examples}, we give a case study with $3$ test triple examples which are inferred by axioms during training. 
Using the third one as an example, 
the test triple is (\texttt{Jenny\_McCarthy},  $/type\_of\_union$,  \texttt{Marriage}). \texttt{Jenny\_McCarthy} is a person with only a few triples in KG and is a sparse entity with sparsity $0.997$. \texttt{Marriage} is a type of union for people which has many links with individual person and its sparsity is $0.607$. The subject prediction rank $(?,\;  $/type\_of\_union$,  \texttt{Marriage})$ in ANALOGY is $4199$ and the object prediction (\texttt{Jenny\_McCarthy},  $/type\_of\_union$,  ?) is $2$, from which we can see that the prediction for a sparse entity is much worse than a frequent entity. In IterE, this triple is predicted by rule (\texttt{X}, $/type\_of\_union$, \texttt{Z}) $\gets$ (\texttt{X}, $spouse$, \texttt{Y}), (\texttt{Y}, $/type\_of\_union$, \texttt{Z}) because there are (\texttt{Jenny\_McCarthy}, $spouse$, \texttt{Jim\_Carrey}) and (\texttt{Jim\_Carrey}, $/type\_of\_union$, \texttt{Marriage}) in KG which can compose a grounding for previous rules. Thus during training, the test triple is inferred by IterE and injected into embedding learning. Thus the subject prediction is improved from $4199$ to $33$ compared with ANALOGY and the object prediction keeps the same. When we predict with both embeddings and axioms (IterE + axioms), the subject prediction is improved from $33$ to $1$ and the object prediction from $2$ to $1$ compared with IterE. Other two cases in Table~\ref{sparse-entity-prediction-examples} perform similar to the third one.

These examples show that:
(1) adding triples that can be inferred by axioms back into training improves the prediction results of related sparse entities without hurting the results for non-sparse entities. 
(2) the deductive ability of axioms helps ensure truth value of triples with sparse entities, which, in our opinion, overcomes uncertainties and noises that effect embedding prediction.

\vspace{-2mm}
\section{Related Work}
\label{sec:related-work}

In this paper, we focus on iteratively learning embeddings and rules from knowledge graphs. 
Thus the related work includes two parts: (1) embedding learning, (2) rule learning. 
\vspace{-1mm}
\subsection{Embedding Learning}
\vspace{-1mm}
Knowledge graph embedding learns latent representations for entities and relations in continuous vector space.
Normally entities are represented as vectors \cite{TransE:conf/nips/BordesUGWY13}\cite{HolE:conf/aaai/NickelRP16}\cite{ComplEx:conf/icml/TrouillonWRGB16} and relations are represented as vectors \cite{TransE:conf/nips/BordesUGWY13}\cite{TransH:conf/aaai/WangZFC14}\cite{HolE:conf/aaai/NickelRP16} or matrices \cite{DistMult:conf/iclr/2015}\cite{RESCAL:conf/icml/NickelTK11}\cite{ANALOGY:conf/icml/LiuWY17}. These embeddings are assumed to preserve the semantics in a knowledge graph such as the similarity between entities\cite{RDF2Vec:conf/semweb/RistoskiP16}, the truth of triples\cite{TransE:conf/nips/BordesUGWY13}. For most embedding methods, the input are triples existing in knowledge graph and embeddings are trained based on different vector space assumptions, such as translation-based assumption in 
\cite{TransE:conf/nips/BordesUGWY13}
\cite{TransH:conf/aaai/WangZFC14}
\cite{TransR:conf/aaai/LinLSLZ15} 
\cite{TransD:conf/acl/JiHXL015}, and linear map assumption in 
\cite{RESCAL:conf/icml/NickelTK11} 
\cite{DistMult:conf/iclr/2015} 
\cite{ComplEx:conf/icml/TrouillonWRGB16}
\cite{ANALOGY:conf/icml/LiuWY17}. 
Some methods do not follow specific vector space assumption for embeddings, and special neural networks are applied to train embeddings and assign reasonable scores for different triples. For example, neural tensor networks in \cite{NTN:conf/nips/SocherCMN13}, convolution neural network in \cite{ConvE:conf/aaai/2018} and shared memory neural network in \cite{IRN:conf/rep4nlp/ShenHCG17}.
Linear map assumption is adopted for embedding learning in this paper because of its simplicity and good property for rule learning, while other assumptions or neural network models don't have.

Besides triples in KGs, some embedding methods also utilize other information during learning, for examples, entity descriptions in \cite{Jointly:conf/emnlp/WangZFC14}\cite{DKRL:conf/aaai/XieLJLS16}\cite{TEKE:conf/ijcai/WangL16}\cite{SSP:conf/aaai/0005HMZ17}, entity types in \cite{SSE:conf/acl/GuoWWWG15}\cite{TKRL:conf/ijcai/XieLS16}, 
entity images in \cite{IKRL:conf/ijcai/XieLLS17},
and  paths in \cite{RTransE:conf/emnlp/Garcia-DuranBU15}\cite{PTransE:conf/emnlp/LinLLSRL15}\cite{CVSM:conf/acl/NeelakantanRM15}, which can be regarded as a kind of rules.

Rules, as useful information for reasoning, are also considered to assist embedding learning.
Among these methods different types of rules are considered, for examples, 
\cite{rules1:conf/ijcai/WangWG15} introduces one type of logical rule and three physical rules, 
\cite{RUGE:conf/aaai/GuoWWWG18} utilizes horn soft rules learned from \cite{AMIE+:journals/vldb/GalarragaTHS15}, 
\cite{PKDD17:conf/pkdd/MinerviniCMNV17} incorporates equivalence and inversion Axioms, and 
\cite{KALE:conf/emnlp/GuoWWWG16} considers two types of logical rules. 
To incorporate rules into embedding learning, some methods try to improve the embedding method's capability of modeling rules by various ways, such as  encoding relations as convex regions in \cite{KR18:conf/kr/Gutierrez-Basulto18} and non-negativity constraints on entity representations in \cite{ACL18:conf/acl/WangWGD18}. 
Some methods regard rules as guidance for embedding learning. For examples, \cite{rules1:conf/ijcai/WangWG15} forms the whole learning process as a integer linear programming problem with conditions defined based on rules. \cite{RUGE:conf/aaai/GuoWWWG18} infers unlabeled triples according to  input rules.  \cite{PTransE:conf/emnlp/LinLLSRL15} adds constrains for relation embeddings participating in path rules. \cite{Lifted:conf/emnlp/DemeesterRR16} maps entity-tuple embedding into an approximately Boolean space and encourages a partial ordering over relation embedding based on implication rules. \cite{ACL18:conf/acl/WangWGD18} adds approximate entailment constraints on relation representations.
Some methods jointly embed rules and triples during learning, such as \cite{KALE:conf/emnlp/GuoWWWG16} which first learns rules based on embeddings from \cite{TransE:conf/nips/BordesUGWY13} and then retrains the embeddings incorporating with rules.  

Aforementioned embedding methods with rules normally make rule learning detached from embedding learning and unrelated with embeddings.  
While our goal in this paper is not only an embedding learning system, but also a rule learning system in which rules are learned from embeddings and help improve embedding quality.

\vspace{-2mm}
\subsection{Rule learning}
Rule is an important component for reasoning and has been studied in many previous works. 
Among rule learning works, different kinds of rules are adopted, for example, Horn rules in \cite{AMIE+:journals/vldb/GalarragaTHS15}, closed path rules in \cite{PRA:conf/emnlp/LaoMC11} and general rules with both variables and constants and atoms occurring either positively or negatively in \cite{RULES:conf/semweb/HoSGKW18}, frequent predicate cycles in \cite{RDF2rule:journals/corr/WangL15i}, semantic association rules in \cite{SWARM:conf/pricai/BaratiBL16}.
These methods all consider the types of rules to learn from their structures, while in this paper, we consider rules based on  their semantics 
because we think semantics are important for the development of knowledge graphs. Thus we adopt OWL2 object property expression axioms to form the types of rules to learn.

To calculate the confidence of rules, standard confidence or PCA confidence are usually used \cite{AMIE+:journals/vldb/GalarragaTHS15}\cite{RULES:conf/semweb/HoSGKW18} which are based on searching for supports of rules in the whole knowledge graph. To reduce the search space, many rule learning systems have a pruning strategy.  
Apart from graph search, there are also some works trying to learn both structures and scores of rules based on deep neural models \cite{NeuralLP:conf/nips/YangYC17} or reinforcement learning \cite{DeepPath:conf/emnlp/XiongHW17}\cite{GoForAWalk:journals/corr/abs-1711-05851}.

Embeddings are also used to help rule learning and are used for different purposes. Some utilize embeddings to guide and prune the search for candidate rules\cite{RLvLR:conf/ijcai/OmranWW18}. Some make embeddings to complete the knowledge graph during rule learning\cite{RULES:conf/semweb/HoSGKW18}. Some use embeddings to assign scores for rules\cite{DistMult:conf/iclr/2015}\cite{KALE:conf/emnlp/GuoWWWG16}. They all rely on calculations among embeddings and the way of calculations depends on specific embedding methods.

Different from these methods which mainly proposed to learning rules, we devote to learn embeddings and rules at the same time and make their advantages contribute to each other's learning.

\vspace{-1mm}
\section{Conclusion and Future Work}
\label{sec:conclusion}
In this paper, we discuss the advantages and disadvantages of two common knowledge graph reasoning methods, embedding-based method and rule-based method, and propose IterE that iteratively 
learns
embeddings and rules in one model and 
enjoys the mutual benefit between them. 
In the future, we will continuously investigate combining inductive and deductive reasoning together 
and develop models that could unify different kinds of reasoning.

\vspace{-2mm}
\begin{acks}
This work is funded by NSFC91846204/61473260, national key research program YS2018YFB140004, Alibaba CangJingGe(Knowledge Engine) Research Plan and SNF Sino Swiss Science and Technology Cooperation Programme program under contract RiC 01-032014.
\end{acks}

\bibliographystyle{ACM-Reference-Format}
\balance 
\bibliography{embedding}

\end{document}